# Hierarchical Evaluation Function: A Multi-Metric Approach for Optimizing Demand Forecasting Models

Adolfo González[1*] Víctor Parada[2*]

[1,2] Department of Computer Engineering and Informatics, Faculty of Engineering, University of Santiago Chile

**Abstract**

Demand forecasting in competitive, uncertain business environments requires models that can integrate multiple evaluation perspectives rather than being restricted to hyperparameter optimization based on a single metric. This traditional approach tends to prioritize one error indicator, which can bias results when metrics provide contradictory signals. In this context, the Hierarchical Evaluation Function (HEF) is proposed as a multi-metric framework for hyperparameter optimization that integrates explanatory power ($R^2$), sensitivity to extreme errors (RMSE), and average accuracy (MAE). The performance of HEF was assessed using four widely recognized benchmark datasets in the forecasting domain: Walmart, M3, M4, and M5. Prediction models were optimized through Grid Search, Particle Swarm Optimization (PSO), and Optuna, and statistical analyses based on difference-of-proportions tests confirmed that HEF delivers superior results compared to a unimetric reference function, regardless of the optimizer employed, with particular relevance for heterogeneous monthly time series (M3) and highly granular daily demand scenarios (M5). The findings demonstrate that HEF improves stability, generalization, and robustness at low computational cost, consolidating its role as a reliable evaluation framework that enhances model selection, enables more accurate demand forecasts, and supports decision-making in dynamic, competitive business environments.

* Corresponding author

1 E-Mail: adolfo.gonzalez.c@usach.cl; https://orcid.org/0009-0008-4452-1285

2 E-Mail:: victor.parada@usach.cl; https://orcid.org/0000-0002-8649-5694

## 1. Introduction

Inventory management in competitive and dynamic business environments constitutes a multidimensional challenge, especially when decisions must be made under conditions of high demand uncertainty, budget constraints, and physical storage limitations. In this context, anticipating demand for multiple products, optimally allocating financial resources, and considering logistics capacities represent an NP-Hard problem due to its combinatorial nature and the concurrence of multiple constraints (Cárdenas-Barrón, Melo, & Santos, 2021; Zhang, Xie, & Sarin, 2021). The selection of inventory units that maximize profitability, considering profit margins, replenishment times, and acquisition costs, is conditioned by the spatial constraints inherent to packaging problems. Under this framework, accurate demand prediction becomes a crucial component of strategic planning, enabling resource optimization, cost reduction, and anticipating market variations (Samal & Ghosh, 2026). Although statistical models have historically been used, the adoption of machine learning techniques has increased in recent years (Wahedi, et al., 2023; Tang, 2024; Tan, Gu, Xu, & Li, 2024; Jahin, Shahriar, & Amin, 2025; Mittal, 2024; Yang, 2025). However, the stochastic nature of demand, influenced by external factors and abrupt structural changes, remains a substantial challenge (Trull, García-Díaz, & Peiró-Signes, 2024; Peláez, Aneiros, & Vilar, 2024; Riachy, et al., 2025).

In high-uncertainty scenarios, forecasting models require flexibility to adapt to abrupt variations in consumption patterns, which demands robust modeling frameworks and rigorous evaluation mechanisms to guide parameter and hyperparameter optimization. These processes often rely on evaluation functions based on metrics such as Mean Absolute Error (MAE) and Root Mean Squared Error (RMSE), which provide complementary insights: MAE is more robust against outliers, while RMSE penalizes extreme errors more severely (Chicco, Warrens, & Jurman, 2021; Hyndman & Athanasopoulos, 2021; Koutsandreas, Spiliotis, Petropoulos, & Assimakopoulos, 2022). However, the isolated use of individual metrics can lead to partial or biased interpretations (Seyedzavvar, Boğa, & Soudmand, 2025). As Koutsandreas et al. (2022) point out, there is no consensus on which metric should prevail; therefore, it is recommended to use multiple indicators together to achieve a more comprehensive evaluation (Samal & Ghosh, 2026).

In this sense, composite evaluation functions have attracted growing interest, as they integrate various metrics under hierarchical or weighted schemes, providing more balanced estimates of model performance (Seyedzavvar, Boğa, & Soudmand, 2025; Wang, Ou, Liu, Du, & Wang, 2025; Samal & Ghosh, 2026; Moradpour, Ritter, & Hauschild, 2025). In demand prediction, selecting an appropriate evaluation function is crucial, as an inappropriate choice can introduce bias and compromise accuracy in real-world environments (Muñoz, Ballester, Copaci, Moreno, & Blanco, 2025; Seiringer, Altendorfer, Felberbauer, Bokor, & Brockmann, 2025; Wang, Ou, Liu, Du, & Wang, 2025). Despite advances in prediction algorithms and heuristic or metaheuristic techniques, challenges related to scalability, computational efficiency, and adaptability to changing contexts persist (Iqbal, Gil, & Kim, 2025).

Given this scenario, the design of custom evaluation functions appears to be an effective alternative for combining metrics in a consistent and flexible manner. However, the lack of clear guidelines for their selection and implementation limits objective comparisons between predictive approaches (Amin, Aladdin, Hasan, Mohammed-Taha, & Rashid, 2023; Vilar, 2025; Seyedzavvar, Boğa, & Soudmand, 2025). To address this gap, this research proposes the Hierarchical Evaluation Function (HEF), a hierarchical and dynamic function that integrates the $R^2$, MAE, and RMSE metrics to efficiently guide the optimization of predictive models in highly variable contexts. The proposal is supported by empirical evidence demonstrating that the combination of metrics improves the discriminatory capacity between models compared to univariate evaluations (Ferouali, Elamrani Abou Elassad, Qassimi, & Abdali, 2025; Pakdel, He, & Chen, 2025). Furthermore, its integration with optimization techniques, such as Grid Search, Particle Swarm Optimization (PSO), and Bayesian optimization, is proposed, aiming to select optimal hyperparameter configurations in a flexible and efficient manner (Muñoz, Ballester, Copaci, Moreno, & Blanco, 2025; Iqbal, Gil, & Kim, 2025; Wang, Ou, Liu, Du, & Wang, 2025).

Within this framework, the study confirms that the choice of evaluation function has a decisive impact on the performance of demand forecasting models. The results show that HEF consistently outperforms the MAE used as the evaluation function in global metrics, such as $R^2$, Global Relative Accuracy (GRA), RMSE, and RMSSE, thereby strengthening its explanatory capacity and robustness against large errors. In contrast, MAE as an evaluation function maintains advantages in absolute error reduction (MAE and MASE) and computational efficiency, making it a practical alternative in short-term or resource-constrained scenarios. Consequently, the choice between the two functions should not be considered mutually exclusive, but rather a strategic decision dependent on the application's objectives: HEF emerges as the most appropriate option for business planning contexts and long-term horizons, while MAE proves more efficient in operational and short-term settings (Samal & Ghosh, 2026). In summary, this study demonstrates that HEF represents a robust and adaptive approach for enhancing the accuracy and stability of demand forecasting models in dynamic environments, thereby overcoming the limitations of one-dimensional metrics.

The article is organized as follows: after the introduction and literature review, the methodological foundations are presented, including the mathematical formulation of the HEF, the forecasting models considered, and the optimizers employed. Subsequently, the experimental design is described, and comparative results against benchmark functions are presented. In the final section, the findings are discussed and conclusions are proposed, along with possible lines of future research.

## 2. Literature Review

Demand forecasting is a central element in strategic planning across various sectors, including industrial production, hospital management, and energy planning. Several studies demonstrate that the ability to anticipate demand enables resource optimization, cost reduction, and greater responsiveness to changes in market dynamics. In the energy sector, short-term electricity demand forecasting has been addressed using seasonal time series models, which have improved projection accuracy and supported decision-making in supply management (Trull, García-Díaz, & Peiró-Signes, 2024), while in the hospital sector, estimating patient flows and bed occupancy based on real-time data has optimized the allocation of medical resources and hospital capacity (García de Vicuña Bilbao, López-Cheda, Jácome, & Mallor Giménez, 2023; Tuominen, et al., 2024). Likewise, in electricity markets, regression models have been applied to predict demand curves, providing robust prediction intervals and increasing forecast reliability (Peláez, Aneiros, & Vilar, 2024). In water resource management, the integration of artificial intelligence techniques has enabled more accurate estimations of water demand, promoting efficient use and reducing environmental impact. Along these lines, Otamendi et al. (2024) show that combining neural networks with seasonality-based preprocessing techniques improves model fitting, although they do not explore customized evaluation functions, which represents an opportunity for future research. Overall, these approaches highlight the importance of advanced methodologies for more efficient, resilient, and sustainable planning (Li, Zhang, Zhang, & Wang, 2024; Riachy, et al., 2025; Nguyen, et al., 2026; Sapnken, et al., 2026; Wei & Mukherjee, 2024; Samal & Ghosh, 2026).

The accuracy of predictive models is crucial to ensure consistency between projections and observed conditions, particularly over long-term horizons, where cumulative errors can increase strategic risks (Parga-Prieto & Aranda-Pinilla, 2018). In this sense, the evolution of optimization techniques has given rise to approaches that integrate learning dynamics, allowing continuous adjustments that reduce cumulative error and strengthen the robustness of forecasts (Babii, Ghysels, & Striaukas, 2022; Iqbal, Gil, & Kim, 2025; Amin, Aladdin, Hasan, Mohammed-Taha, & Rashid, 2023). However, demand prediction models still face limitations related to scalability, computational efficiency, and adaptability to changing environments (Muñoz, Ballester, Copaci, Moreno, & Blanco, 2025), which has motivated the design of customized evaluation functions as a promising strategy to optimize performance, facilitate efficient hyperparameter selection, and improve generalization capacity. Examples include the combination of XGBoost with specific evaluation functions, which has enabled tuning key parameters in high-variability scenarios (Pandove & Deepika., 2025), and the optimization of models in embedded systems, which has reported improvements in accuracy and efficiency, as highlighted by Touzout et al. (2025), who apply adaptive heuristics that, although they do not include explicit evaluation functions, lay the groundwork for their integration (Hong, Bai, Huang, & Chen, 2024; Samal & Ghosh, 2026; Seyedzavvar, Boğa, & Soudmand, 2025).

Hyperparameter optimization has emerged as a key strategy for improving predictive accuracy (Muñoz, Ballester, Copaci, Moreno, & Blanco, 2025), particularly through the dynamic tuning of evaluation functions tailored to the changing nature of the data. In this area, Bayesian Optimization has demonstrated effectiveness in calibrating energy prediction models by enabling adaptive metric selection without significantly increasing computational costs (Iqbal, Gil, & Kim, 2025; Muñoz, Ballester, Copaci, Moreno, & Blanco, 2025). Meanwhile, the combination of neural networks with weighted error functions has been shown to mitigate the impact of outliers and improve training stability (Pakdel, He, & Chen, 2025). Additionally, metaheuristic techniques such as Particle Swarm Optimization (PSO) (Talbi, 2009) have proven effective in dynamically modifying hyperparameters during iterative optimization processes; in particular, Khan et al. (2024) document the use of PSO with weighted MAE-based objective functions, adjusting penalties based on environmental volatility. However, inappropriate configurations can increase computational cost without yielding substantial improvements (Muñoz, Ballester, Copaci, Moreno, & Blanco, 2025), which has motivated recent proposals focused on adaptive approaches that balance accuracy and efficiency, thereby reducing error without compromising inference speed (Babii, Ghysels, & Striaukas, 2022; Muñoz, Ballester, Copaci, Moreno, & Blanco, 2025). In this context, metaheuristic methods have demonstrated great versatility in optimizing complex and nonlinear models, both in industrial applications and in demand prediction (Hu & Li, 2023; Muñoz, Ballester, Copaci, Moreno, & Blanco, 2025). However, their effectiveness depends on the parameter configuration and the evaluation function guiding the search for optimal solutions. In AutoML environments, Arnold et al. (2024) and Hernández et al. (2024) emphasize that the use of configurable evaluation functions is crucial for avoiding overfitting and enhancing robustness, as they enable the incorporation of domain-specific constraints through dynamic adjustments in real-time (Liu, Cao, Wang, & Xiang, 2025; Moradpour, Ritter, & Hauschild, 2025; Wang, Ou, Liu, Du, & Wang, 2025).

The selection or construction of evaluation functions still lacks systematic criteria, which limits their impact on model accuracy and robustness (Vargas-Forero, Manotas-Duque, & Trujillo, 2024; Muñoz, Ballester, Copaci, Moreno, & Blanco, 2025). While metrics such as the coefficient of determination ($R^2$) and the mean absolute error (MAE) are frequently used (Chicco, Warrens, & Jurman, 2021; Ferouali, Elamrani Abou Elassad, Qassimi, & Abdali, 2025), recent studies indicate that $R^2$ measures the proportion of explained variance, while MAE quantifies the error in the original units by uniformly penalizing deviations (Iqbal, Gil, & Kim, 2025); in this sense, no metric is universally superior and its effectiveness depends on the context and business objectives. Therefore, a combined approach is recommended to achieve more balanced evaluations, as proposed by Zarma et al. (2025), who integrate MAE and Mean Absolute Percentage Error (MAPE) through adaptive weighting for model evaluation in the electricity sector. However, the design of custom evaluation functions remains a challenge in time series modeling, since, although traditional metrics such as $R^2$, MAE, and Mean Squared Error (MSE) are still widely used, their ability to reflect the complexity and adaptability of models is limited in highly dynamic scenarios (Maturana, et al., 2012; Santiago, Dorronsoro, Fraire, & Ruiz, 2021; He & Aranha, 2024; Bischl, et al., 2023; Arnold, Biedebach, Küpfer, & Neunhoeffer, 2024; Muñoz, Ballester, Copaci, Moreno, & Blanco, 2025). This limitation has driven the exploration of functions that seek to balance accuracy, stability, and adaptability, integrating metrics into hierarchical and weighted structures (Michelucci & Venturini, 2023; Muñoz, Ballester, Copaci, Moreno, & Blanco, 2025), while the incorporation of Bayesian Optimization and metaheuristic techniques has enabled the automation of metric selection, reducing dependence on a single function and improving model robustness (Iqbal, Gil, & Kim, 2025; Pandove & Deepika., 2025; Muñoz, Ballester, Copaci, Moreno, & Blanco, 2025; Buzpinar, et al., 2025; Moradpour, Ritter, & Hauschild, 2025; Sapnken, et al., 2026).

Despite these advances, the definition and validation of adaptive functions remain a challenge, particularly in terms of integrating them with metrics that capture data variability and complexity. Exploring new strategies for designing evaluation functions is, therefore, crucial for the development of artificial intelligence systems used in demand forecasting. Within this framework, this research proposes a hierarchical multi-objective function based on $R^2$, MAE, and RMSE, which, when integrated with algorithms such as PSO, Bayesian Optimization, and Grid Search, aims to overcome the limitations identified in the literature, fostering the development of more adaptive, accurate, and generalizable models.

Studies have increasingly acknowledged the limitations of evaluation functions based on a single metric, which has driven the development of composite or multivariate structures for the assessment and optimization of predictive models (Samantaray, Mooney, & Vivacqua, 2024; Filho, Pais, & Pais, 2025; Samal & Ghosh, 2026). These approaches typically integrate metrics through linear weighting schemes, adaptive scoring rules, or objective functions tailored to specific optimizers, particularly in the context of AutoML and hyperparameter optimization (Tonati, Vece, Giannotti, & Pellungrini, 2025; Hong, Bai, Huang, & Chen, 2024; Liu, Cao, Wang, & Xiang, 2025; Seyedzavvar, Boğa, & Soudmand, 2025; Wang, Ou, Liu, Du, & Wang, 2025). Such strategies have demonstrated greater robustness than univariate criteria, especially in complex, high-dimensional search spaces (Samantaray, Mooney, & Vivacqua, 2024; Filho, Pais, & Pais, 2025; Samal & Ghosh, 2026; Wang, Ou, Liu, Du, & Wang, 2025). However, most existing composite structures exhibit notable limitations. First, they often rely on flat weighting schemes that do not explicitly prioritize metrics based on their operational relevance (Tonati, Vece, Giannotti, & Pellungrini, 2025; Seyedzavvar, Boğa, & Soudmand, 2025). Second, they lack penalization mechanisms to handle logically invalid predictions or extreme errors, which are common in volatile demand scenarios. Finally, these functions often lack interpretability, making it difficult to justify model selection in business contexts (Tonati, Vece, Giannotti, & Pellungrini, 2025; Wang, Ou, Liu, Du, & Wang, 2025). Consequently, a gap remains in the design of evaluation functions that integrate multiple metrics through a hierarchical and penalty-based structure—capable of balancing explanatory power, average precision, and resilience to extreme errors, while preserving interpretability and computational efficiency. The Hierarchical Evaluation Function (HEF) proposed in this work directly addresses this gap.

## 3. Materials and Methods

The optimization of parameters and hyperparameters in prediction models is often addressed as a one-dimensional problem, relying on functions based on metrics such as MAE or RMSE (Seyedzavvar, Boğa, & Soudmand, 2025). Although these metrics provide relevant insights, they are insufficient for a comprehensive evaluation of performance (Michelucci & Venturini, 2023; Samal & Ghosh, 2026; Wang, Ou, Liu, Du, & Wang, 2025). Likewise, explanatory metrics such as $R^2$ tend to overestimate model fit in scenarios with high error dispersion (Bischl, et al., 2023; Seyedzavvar, Boğa, & Soudmand, 2025). These limitations underscore the need for an approach that integrates complementary metrics within a hierarchical framework, allowing for the simultaneous evaluation of accuracy, robustness, and explanatory power (Wang, Ou, Liu, Du, & Wang, 2025). In this context, the Hierarchical Evaluation Function (HEF) is proposed, integrating $R^2$, MAE, and RMSE through adaptive weighting and penalties, thereby providing a unified framework to guide the selection and optimization of demand forecasting models.

### *3.1. Hierarchical Evaluation Function (HEF)*

The Hierarchical Evaluation Function (HEF) is based on three core metrics, $R^2$, MAE, and RMSE, selected for their complementarity, numerical stability, and operational interpretability in volatile forecasting scenarios, as justified throughout this section. This selection aligns with recommendations in the literature, which emphasize $R^2$ for global explanatory power, MAE for stable and interpretable average error, and RMSE for penalizing extreme deviations (Chicco, Warrens, & Jurman, 2021; Michelucci & Venturini, 2023; Hajiabbasi Somehsaraie, Tofighi, Wang, Wang, & Xiao, 2025). In practice, the tuning of parameters and hyperparameters is often formulated as an optimization problem of a single error metric, where $y$ denotes the observed values, $\hat{y}_\theta$ the predictions of the model parameterized by $\theta$, and $\mathcal{L}$ a one-dimensional loss function.

(1)

$$\hat{\theta} = \arg \min_{\theta \in \Theta} \mathcal{L}(y, \hat{y}_\theta)$$

Typically, $\mathcal{L}$ denotes the Mean Absolute Error (MAE) or the Root Mean Squared Error (RMSE). While these metrics provide complementary insights, MAE is more robust to outliers, whereas RMSE penalizes large errors more severely. However, their isolated use may introduce bias and compromise the model's generalizability in volatile scenarios (Chicco, Warrens, & Jurman, 2021; Koutsandreas, Spiliotis, Petropoulos, & Assimakopoulos, 2022; Hodson, 2022; Iacopini, Ravazzolo, & Rossini, 2023). Similarly, the coefficient of determination ($R^2$) quantifies the explanatory power of the model but may overestimate the fit in cases of high error dispersion (Ferouali, Elamrani Abou Elassad, Qassimi, & Abdali, 2025). In these definitions, $y_t$ denotes the actual value at time t, $\hat{y}_t$ the predicted value at time t, $\bar{y}_t$ the average of the actual values, and n the total number of observations. The mathematical expressions are defined as:

$$MAE = \frac{1}{n} \sum_{t=1}^{n} |y_t - \hat{y}_t| \tag{2}$$

$$RMSE = \sqrt{\frac{1}{n} \sum_{t=1}^{n} (y_t - \hat{y}_t)^2} \tag{3}$$

$$R^2 = 1 - \frac{\sum_{t=1}^{n} (y_t - \hat{y}_t)^2}{\sum_{t=1}^{n} (y_t - \bar{y}_t)^2} \tag{4}$$

In mathematical terms, the problem is formulated through a hierarchical evaluation function that integrates multiple metrics, aiming to balance accuracy, robustness, and explanatory power. Although the design of HEF resembles a multi-objective scheme by combining these dimensions, it is not a Pareto-based optimization. Instead, a hierarchical structure was adopted to prioritize metrics sequentially and reduce the computational complexity associated with formal multi-objective approaches. Formally, the set of observed values is denoted as $y$, the predictions generated by the model parameterized by $\theta$ are represented as $\hat{y}_\theta$, and $\theta \in \Theta$ corresponds to the vector of parameters or hyperparameters of the model. Under this formulation, the goal is to reduce both mean and extreme errors while simultaneously maximizing the explanatory capacity of the model. This idea can be expressed as:

$$\min_{\theta \in \Theta} (MAE(y, \hat{y}_\theta), RMSE(y, \hat{y}_\theta), 1 - R^2(y, \hat{y}_\theta)) \tag{5}$$

This approach does not constitute a multi-objective problem in the strict sense of Pareto optimization, but rather reflects the need to integrate complementary metrics into a single, composite, and hierarchical evaluation function. The absence of such a unified criterion in traditional approaches can hinder the objective selection of the best model (Muñoz, Ballester, Copaci, Moreno, & Blanco, 2025; Zarma, et al., 2025), thus justifying the proposal of the Hierarchical Evaluation Function (HEF).

In this formulation, $R^2$, MAE, and RMSE are normalized by the mean of the actual training values ($\bar{y}$). The coefficients $\omega_{R^2}$ and $\omega_{MAE}$, $\omega_{RMSE}$ represent the hierarchical weights assigned to each evaluation component, while $\Pi(\hat{y})$ introduces penalties in the case of negative or inconsistent predictions. This framework enables the integration of accuracy, robustness, and explainability within a single structure, overcoming the fragmentation of traditional approaches (Hernandez, Saini, & Moore, 2024; Khan, et al., 2024). With these considerations, the HEF is defined as:

$$\mathcal{F}_{HEF}(\theta) = \omega_{R^2}(1 - R^2) + \omega_{MAE}\left(\frac{MAE}{\bar{y}}\right) + \omega_{RMSE}\left(\frac{RMSE}{\bar{y}}\right) + \Pi(\hat{y}) \tag{6}$$

To reduce the scale dependency inherent in absolute error metrics, the MAE and RMSE are normalized by the mean of the training series (Hyndman & Koehler, 2006). This strategy expresses errors relative to the average demand level, facilitating comparisons between models trained on series with different magnitudes (Makridakis, Spiliotis, & Assimakopoulos, 2022). This approach is conceptually consistent with the use of scaled metrics, such as MASE or RMSSE, widely adopted in forecasting studies and competitions, where normalization aims to ensure comparability without distorting the semantic meaning of the error (Makridakis, Spiliotis, & Assimakopoulos, 2022; Hyndman & Koehler, 2006; Khan & Osińska, 2023). In this context, normalization by the mean does not seek to fully homogenize the metrics, but rather to remove scale effects while preserving the interpretation of each indicator within the hierarchical evaluation function (Athanasopoulos & Kourentzes, 2023).

More comprehensively, the mathematical formulation of the HEF is expressed as a case-dependent system of tolerance thresholds and validity conditions. In this formulation, $\Pi_k$ represents progressive penalties applied when tolerance thresholds are exceeded, while $\Pi_{invalid}$ establishes a severe penalty in the case of negative or inconsistent predictions. This design ensures that the evaluation is robust to incomplete or invalid values and that it dynamically adapts to the type of metric and the optimization strategy employed (Amin, Aladdin, Hasan, Mohammed-Taha, & Rashid, 2023; Babii, Ghysels, & Striaukas, 2022), guaranteeing not only local accuracy but also stability and logical consistency in real-world demand forecasting scenarios.

It should be noted that, even after normalization, the metrics integrated into the HEF may operate on different ranges, which represents an intentional design decision (Hyndman & Koehler, 2006). Rather than enforcing full homogenization, the hierarchical function preserves the semantic differences between goodness-of-fit indicators ($R^2$) and error-based metrics (MAE and RMSE), managing their relative influence through differentiated weights and progressive penalties (Chai & Draxler, 2014; Koutsandreas, Spiliotis, Petropoulos, & Assimakopoulos, 2022). This approach has been recommended in the literature as an effective alternative for capturing complementary dimensions of predictive performance, preventing a single metric from dominating the optimization process (Makridakis, Spiliotis, & Assimakopoulos, 2018; Deb, Pratap, Agarwal, & Meyarivan, 2002).

$$f(\theta) = \begin{cases} min\left\{\omega_{R^2}(1-R^2) + \omega_{MAE} \cdot \dfrac{MAE}{\bar{y}} + \omega_{RMSE} \cdot \dfrac{RMSE}{\bar{y}}\right\}, & if\ MAE, RMSE\ \leq\ tolerance\ thresholds \\ \Pi_k \left[\omega_{R^2}(1-R^2) + \omega_{MAE} \cdot \dfrac{MAE}{\bar{y}} + \omega_{RMSE} \cdot \dfrac{RMSE}{\bar{y}}\right], & if\ MAE\ or\ RMSE\ exceed\ thresholds \\ \Pi_{invalid}, & if\ there\ are\ negative\ or\ invalid\ predictions \end{cases} \tag{7}$$

The evaluation function $f(\theta)$ responds to the need to balance accuracy and explanatory power in a robust manner across different demand forecasting contexts. Recent studies have shown that the isolated use of metrics such as $R^2$ can lead to an overestimation of performance, particularly in cases with high error dispersion (Chicco, Warrens, & Jurman, 2021). On the other hand, error metrics such as MAE or RMSE, although more sensitive to the magnitude of the error, do not adequately capture the model's ability to explain the variability of the series. Integrating these three metrics into a hierarchical function enables the simultaneous evaluation of both the quality of fit and the impact of extreme errors. This has been empirically supported by works such as Ferouali et al. (2025) and Koutsandreas et al. (2022), which demonstrate that the combination of $R^2$, MAE, and RMSE improves the discrimination between competing models. The inclusion of a penalty for the lack of valid predictions further reinforces the robustness of the approach, preventing models with logical errors or out-of-domain predictions from being considered in the optimization process. Overall, this formulation offers a practical and adaptable solution for scenarios characterized by high uncertainty and volatility, which are typical of demand forecasting in real-world environments.

Although percentage-based error metrics, such as MAPE and its symmetric variant (sMAPE), were foundational in forecasting benchmarks, specifically sMAPE was explicitly used in the M3 competition (Makridakis & Hibon, The M3-Competition: results, conclusions and implications, 2000) and also featured as a primary metric in M4 (Makridakis, Spiliotis, & Assimakopoulos, The M4 Competition: Results, findings, conclusion and way forward, 2018); they were deliberately excluded from the HEF formulation. This decision is based on their well-documented limitations, including numerical instability when actual values are near zero, scale sensitivity across heterogeneous datasets, and asymmetric penalization (Hyndman & Koehler, 2006; Goodwin & Lawton, 1999; Makridakis, Accuracy measures: theoretical and practical concerns, 1993). HEF, instead, integrates $R^2$, MAE, and RMSE for evaluation, while validation metrics such as RMSSE, MASE, and GRA ensure robustness without the drawbacks of percentage-based errors.

In volatile, non-stationary demand environments, the use of individual metrics can lead to biased interpretations, as each captures only a single dimension of model performance (Samal & Ghosh, 2026; Seyedzavvar, Boğa, & Soudmand, 2025). Several studies have emphasized that there is no universally suitable metric for all forecasting contexts (Hyndman & Koehler, 2006; Armstrong & Collopy, 1992; Michelucci & Venturini, 2023; Samal & Ghosh, 2026; Wang, Ou, Liu, Du, & Wang, 2025). For example, $R^2$ evaluates explanatory power, MAE reflects the average error in actual units, and RMSE penalizes extreme errors more severely (Willmott, Matsuura, & Robeson, 2009; Chai & Draxler, 2014; Seyedzavvar, Boğa, & Soudmand, 2025). However, their isolated use may lead to suboptimal decisions, particularly in the presence of high dispersion or structural changes (Badulescu, Hameri, & Cheikhrouhou, 2021; Riachy, et al., 2025). Likewise, percentage-based metrics such as MAPE and sMAPE, though frequently used in benchmarks, have well-documented limitations when applied to series with near-zero values or intermittent demand, often resulting in numerical instability and asymmetric penalization (Goodwin & Lawton, 1999; Kim & Kim, 2016; Hyndman & Koehler, 2006; Samal & Ghosh, 2026). Table 1 summarizes the main strengths, limitations, and risks of adopting these metrics as the sole evaluation criteria, underscoring the need for multivariate or composite approaches to achieve a more balanced and robust assessment.

**Table 1**. Strengths, Limitations, and Risks of Using Individual Forecasting Metrics in Volatile Demand Scenarios.

| Metric | Strengths | Limitations in Volatile Demand | Risks of Using as Sole Metric |
|---|---|---|---|
| $R^2$ | Measures explanatory power and the proportion of variance explained by the model (Hyndman & Koehler, 2006; Seyedzavvar, Boğa, & Soudmand, 2025) | May overestimate performance in series with high dispersion or heteroscedasticity; does not reflect error magnitude (Armstrong & Collopy, 1992; Seyedzavvar, Boğa, & Soudmand, 2025) | Selection of models with good global fit but operational errors |
| MAE | Easy to interpret in original units; robust to outliers; stable during optimization (Willmott, Matsuura, & Robeson, 2009; Samal & Ghosh, 2026) | Treats all errors equally; does not penalize extreme deviations strongly enough (Chai & Draxler, 2014; Seyedzavvar, Boğa, & Soudmand, 2025) | Models with low average error but severe error spikes that impact inventory |
| RMSE | Strongly penalizes large errors; sensitive to extreme deviations (Chai & Draxler, 2014; Seyedzavvar, Boğa, & Soudmand, 2025) | Highly sensitive to outliers; may bias optimization toward overly conservative models (Willmott, Matsuura, & Robeson, 2009) | Overpenalization of rare events and degradation of average error |
| MAPE | Scale-independent; intuitive percentage interpretation (Hyndman & Koehler, 2006; Samal & Ghosh, 2026) | Numerical instability near zero values; biased toward low-demand series (Kim & Kim, 2016) | Distorted evaluation in intermittent demand or near-zero cases |
| sMAPE | Bounded and comparable across series; widely used in benchmarks (Makridakis, Spiliotis, & Assimakopoulos, The M4 Competition: Results, findings, conclusion and way forward, 2018; Samal & Ghosh, 2026) | Asymmetric penalization between over- and underestimation; sensitive to fluctuations (Goodwin & Lawton, 1999) | Tendency to favor models that overestimate demand |
| Single metric (general) | Computationally and conceptually simple | Ignores complementary performance dimensions (Armstrong & Collopy, 1992) | Biased model selection and low robustness in non-stationary environments (Badulescu, Hameri, & Cheikhrouhou, 2021; Riachy, et al., 2025) |

### *3.2. Implementation of the HEF*

The configuration for implementing the HEF is illustrated in Figure 1. This function integrates three complementary metrics $R^2$, MAE, and RMSE, with the objective of balancing the model's explanatory power and predictive accuracy, recognizing that no single metric is universally optimal across all contexts (Chicco, Warrens, & Jurman, 2021; Koutsandreas, Spiliotis, Petropoulos, & Assimakopoulos, 2022; Ferouali, Elamrani Abou Elassad, Qassimi, & Abdali, 2025).

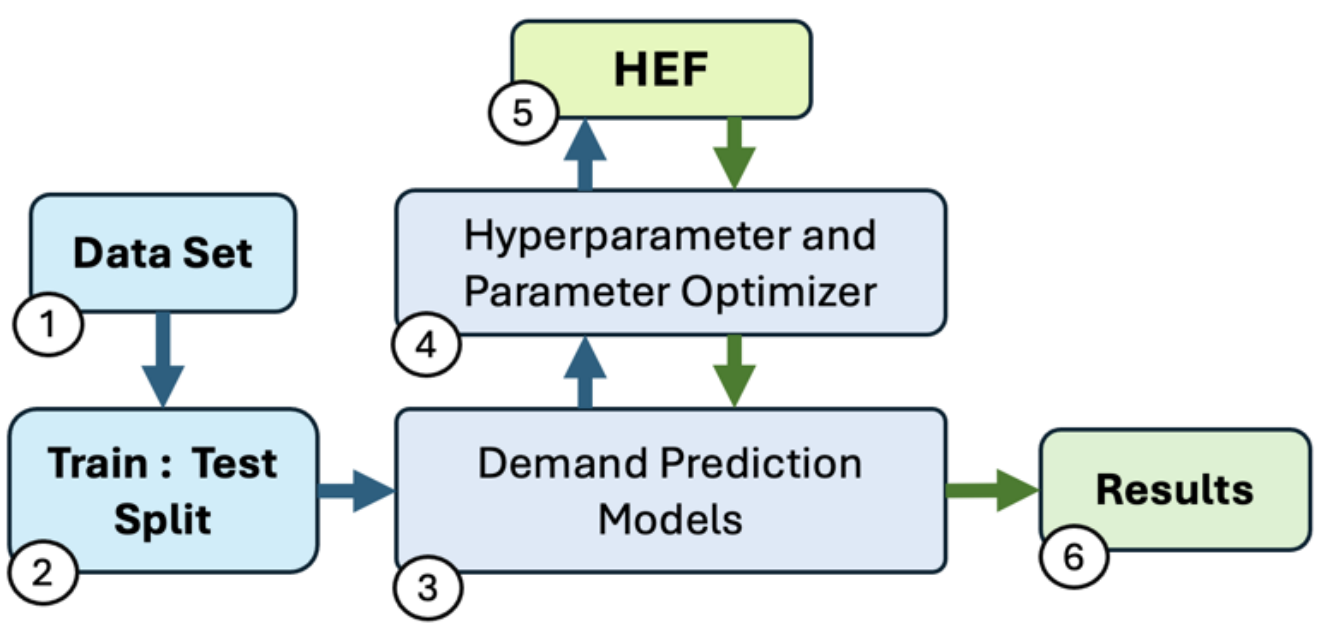

**Figure 1.** The configuration for implementing the HEF.

The process begins with the preparation of the demand dataset (1), which serves as the foundation for the forecast. The data are then divided into training and test subsets (2), which are used during the modeling phase. The demand forecasting models (3) employ the training data for initial calibration and are evaluated on the test data to estimate their generalization capabilities.

In the next stage, a parameter and hyperparameter optimization module (4) is incorporated, responsible for adjusting model configurations according to the complexity of the search space. Different strategies are considered in this context, including exhaustive grid search for finite and discrete configurations, as well as more advanced methods such as Particle Swarm Optimization (PSO) (Talbi, 2009) and Bayesian Optimization using Optuna (Muñoz, Ballester, Copaci, Moreno, & Blanco, 2025; Iqbal, Gil, & Kim, 2025) for continuous or high-dimensional scenarios.

The optimization process is guided by the HEF (5), which serves as the central evaluation function to direct the exploration of the solution space and determine the most suitable configurations. Finally, the adjusted models generate the results (6), which represent the performance achieved under the criteria integrated by the HEF.

### *3.3. Experimental Design*

To evaluate the effectiveness of HEF, an experimental framework was designed to compare its performance with that of another evaluation function under controlled conditions. The objective was to determine whether HEF provides consistent improvements by balancing predictive accuracy and explanatory power across different forecasting models.

Figure 2 illustrates the experimental process, highlighting the main stages, including dataset preparation, model training, HEF-guided hyperparameter optimization, and subsequent results evaluation. This design ensures the systematic validation of the proposed approach and establishes a clear basis for comparison with benchmark functions.

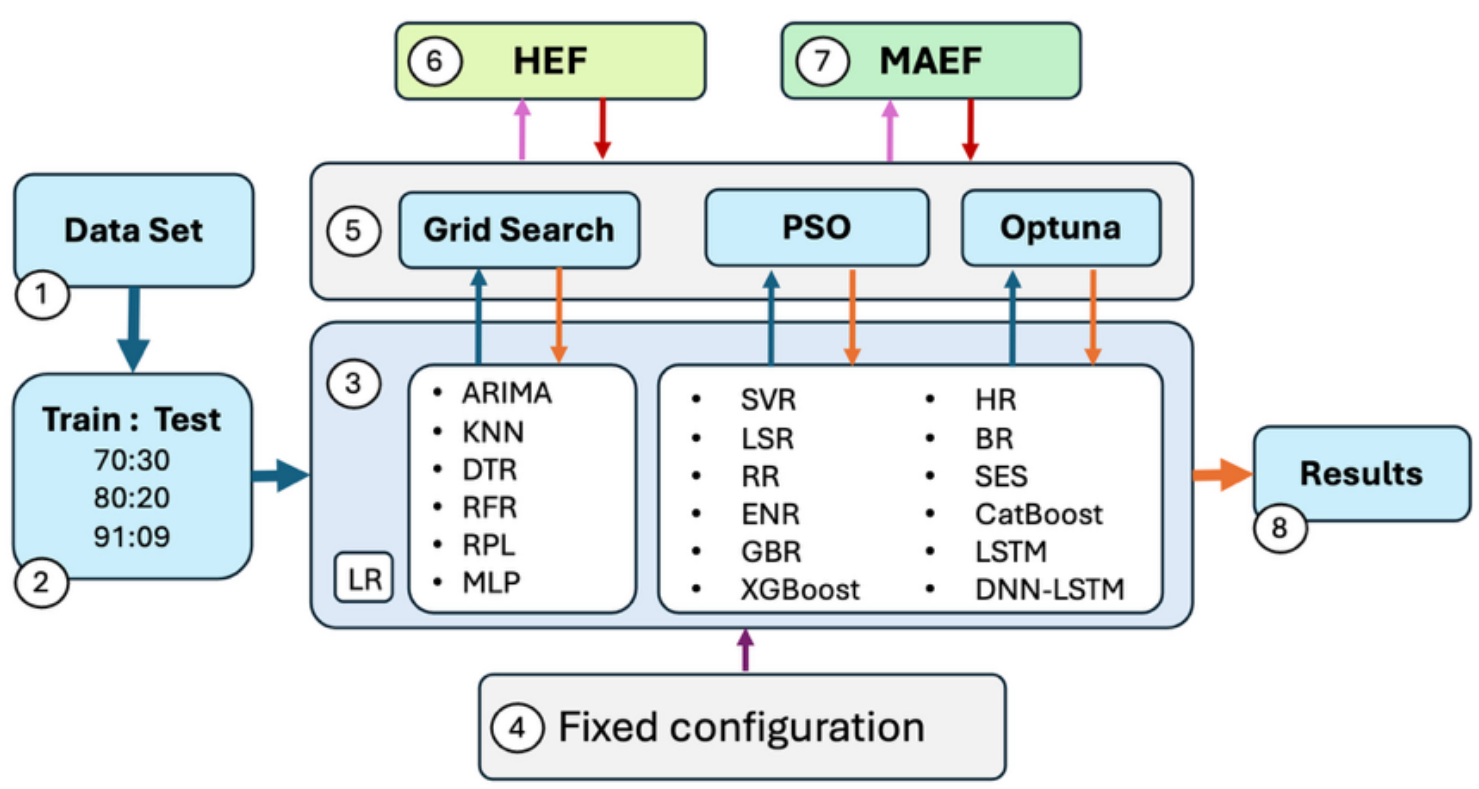

**Figure 2.** Experimental Framework.

#### 3.3.1. Stage 1: Dataset

At this stage, the selected datasets for the experiments were processed and configured. The information used is sourced from various databases and has been treated according to the specific characteristics of each. Table 2 summarizes the datasets used in the study, indicating the number of time series, the total number of observations, and the recording frequency. Among the sources used were Walmart (Yasser, 2021), M3 (Makridakis & Hibon, The M3-Competition: results, conclusions and implications, 2000), M4 (Makridakis, Spiliotis, & Assimakopoulos, 2018), and M5 (Makridakis, Spiliotis, & Assimakopoulos, 2022). For dataset M5, the last n months were selected, and two successive filters were applied: the first eliminated products with no sales in the final part of the analyzed period, and the second discarded those with a high proportion of days without sales throughout the entire series. For datasets M3 and M4, rows and columns with missing values were eliminated, after which the last n periods (weeks or months, depending on the corresponding frequency) were extracted to construct complete and continuous product-level series.

**Table 2.** Datasets used in the experiments.

| Dataset | Frequency | # Time Series | # Observations | Annual Frequency |
|---|---|---|---|---|
| Walmart | Weekly | 46 | 64.35 | 52 |
| M5 | Diary | 1.454 | 1.054.150 | 365 |
| M4 | Weekly | 294 | 28.224 | 52 |
| M3 | Monthly | 1.428 | 34.272 | 12 |
| **Total** | **-** | **3.222** | **1.116.646** | **-** |

Since the experimentation involved multiple iterations, a stratified random sample was drawn from the selected database to ensure statistical representativeness with respect to the full dataset, thereby guaranteeing the validity and generalizability of the results. Stratification was defined according to criteria appropriate to the problem domain, preserving the heterogeneity of the original population in the subsets used for modeling. The sample size was calculated considering finite populations, using the following parameters: the total number of series N in each dataset, a 99% confidence level, a 5% margin of error, and an expected proportion of p = 0.5, which ensures a conservative and robust estimate. The characterization of the extracted data is presented in Table 3.

The selected datasets, Walmart, M3, M4, and M5, were chosen due to their documented non-stationary characteristics, which pose significant challenges for model generalization (Makridakis & Hibon, 2000; Makridakis, Spiliotis, & Assimakopoulos, 2018; Makridakis, Spiliotis, & Assimakopoulos, 2022; Salinas, Flunkert, Gasthaus, & Januschowski, 2020). In particular, Walmart exhibits irregular demand patterns with high intermittency and sudden spikes; M3 contains monthly series with structural breaks and seasonality; M4 comprises weekly series with diverse cycles and long-term trends; and M5 presents highly granular daily data with zero inflation, sporadic bursts, and promotional effects. These properties reflect distinct forms of non-stationarity, such as trend shifts, heteroscedasticity, and temporal irregularity. While this provides a robust empirical basis for testing HEF under volatile conditions, we acknowledge the need to extend validation across other domains (e.g., energy, healthcare, logistics), as discussed in Section 6.4.

**Table 3.** Randomly stratified data set.

| Dataset | Frequency | # Time Series | # Observations | Annual Frequency |
|---|---|---|---|---|
| Walmart | Weekly | 46 | 6435 | 52 |
| M5 | Diary | 650 | 471.250 | 365 |
| M4 | Weekly | 204 | 19.584 | 52 |
| M3 | Monthly | 454 | 10.896 | 12 |
| **Total** | **-** | **1.354** | **508.165** | **-** |

3.3.2. Stage 2: Training and Test Separation

The empirical evaluation was conducted using the Walmart, M5, M4, and M3 datasets, each divided into three training and test configurations (91:9, 80:20, and 70:30). The 91:9 ratio was established with fixed parameters and hyperparameters reported in the literature (Table A1) to ensure replicability with respect to the baseline, while the other partitions allowed for assessing the robustness and generalization capacity of performance under different evaluation functions.

#### 3.3.3. Stage 3: Demand Prediction Models

The prediction models employed range from traditional statistical approaches to modern machine learning techniques and deep neural networks, enabling a comprehensive evaluation of methods with different modeling capabilities:

- ARIMA (AutoRegressive Integrated Moving Average): A classical time series approach that captures linear relationships using autoregressive components, moving averages, and differencing to address trends. It is particularly effective with stationary or stationary-transformable data (Box, Jenkins, Reinsel, & Ljung, 2016).
- K-Nearest Neighbors (KNN): A nonparametric technique that estimates values based on the average of the k nearest neighbors in the feature space, notable for its simplicity, interpretability, and ability to handle nonlinear data (Altman, 1992).
- Support Vector Regression (SVR): An extension of support vector machines for regression, designed to find a function within a tolerance margin of the actual value, using kernel functions to capture complex relationships (Smola & Schölkopf, 2004).
- Linear Regression (LR): A basic regression model that estimates the relationship between independent and dependent variables by minimizing squared error (Montgomery, Peck, & Vining, 2021).
- Lasso Regression (LSR): Incorporates an L1 penalty, promoting automatic variable selection by shrinking less relevant coefficients to zero (Tibshirani, 1996).
- Ridge Regression (RR): Applies an L2 penalty, which mitigates multicollinearity without eliminating variables (Hoerl & Kennard, 1970).
- ElasticNet Regression (ENR): Combines L1 and L2 penalties, allowing the selection of relevant variables while controlling collinearity (Zou & Hastie, 2005).
- Decision Tree Regression (DTR): Produces partitions using hierarchical rules, resulting in interpretable models, though prone to overfitting (Breiman, Friedman, Olshen, & Stone, 1984).
- Gradient Boosting Regressor (GBR): Sequentially builds trees where each new tree corrects the errors of the previous ones, achieving high performance but with greater calibration complexity (Friedman, 2001).
- Random Forest Regressor (RFR): Constructs an ensemble of multiple decision trees trained on random subsets, improving accuracy and reducing overfitting (Breiman, Random forests, 2001).
- XGBoost: An optimized version of gradient boosting that enhances speed, regularization, and missing data handling, making it particularly effective for structured problems (Chen & Guestrin, 2016).
- Huber Regressor (HR): A robust technique that employs the Huber loss function, combining sensitivity to small squared errors with resistance to outliers (Huber, 1992).
- Polynomial Regression (PLR): Extends linear regression with polynomial terms, enabling straightforward modeling of nonlinear relationships (Montgomery, Peck, & Vining, 2021).
- Bayesian Ridge Regression (BR): Introduces prior distributions over coefficients, incorporating uncertainty levels into the estimates (MacKay, 1992).

- Simple Exponential Smoothing (SES): Applied to time series without trend or seasonality, assigning greater weight to recent observations through an exponential decay rule (Brown, 1959).
- CatBoost: A boosting algorithm that efficiently handles categorical variables without explicit encoding and reduces overfitting through ordered boosting (Prokhorenkova, Gusev, Vorobev, Dorogush, & Gulin, 2018).
- Multilayer Perceptron (MLP): An artificial neural network with one or more hidden layers, capable of capturing complex nonlinear relationships and widely applied in regression and classification tasks (Haykin, 1994).
- LSTM (Long Short-Term Memory): A recurrent architecture designed to capture long-term dependencies in temporal sequences, particularly useful in time series (Hochreiter, 1997). In this study, a regression-oriented variant is implemented, comprising two LSTM layers (with 128 and 64 units, respectively, and tanh activation), a 30% dropout mechanism, an intermediate dense layer with 32 neurons (ReLU activation), and a linear output layer. The model is compiled with the Adam optimizer and uses MSE as the loss function.
- DNN-LSTM: A hybrid architecture combining deep dense networks (DNNs) for feature extraction with LSTMs for modeling temporal dynamics, offering robustness in predictions with complex sequential patterns (Goodfellow, Bengio, Courville, & Bengio, 2016). The sequential input is processed by two LSTM layers (128 and 64 units, tanh activation, 30% dropout), while the exogenous input is processed in a three-layer dense network (64, 32, and 16 neurons, ReLU). Both outputs are concatenated and passed through additional dense layers (with 64 and 32 neurons, respectively, and ReLU activation) before a linear output layer is applied. The model is compiled with Adam and uses MSE as the loss function.

The models are classified into two groups: Exhaustive Search (ES), which performs exhaustive exploration over discrete and finite spaces (ARIMA, KNN, DTR, RFR, PLR, and MLP); and Search in Continuous Space (SCS), which includes parameterized models optimized through hyperparameter search techniques in high-dimensional continuous spaces (SVR, LSR, RR, ENR, GBR, XGBoost, HR, BR, SES, CatBoost, LSTM, and DNN-LSTM). This classification categorizes the groups to which optimization techniques are applied, based on the specifications of each model.

#### 3.3.4. Stage 4: Fixed configuration

The definition of fixed parameters for the selected models is based on configurations used in previous research on a subset of the datasets employed. This strategy establishes the baseline for comparing the results obtained with other configurations or optimizations, as reported by Yasser (2021), Ahmedov (2021), Bharat (2021), and Mostafa (2021). The complete set of fixed parameters and hyperparameters is provided in Appendix A (Table A1).

#### 3.3.5. Stage 5: Optimizers

Three complementary approaches were used to optimize the parameters and hyperparameters of the models described:

- Grid Search: Performs an exhaustive search of all possible combinations within a discrete, finite space. It guarantees the identification of the best configuration in the defined domain, although it presents a high computational cost in large or multidimensional spaces (Bergstra & Bengio, 2012). In this study, it was applied to the group of models known as Exhaustive Search (ES).

- Particle Swarm Optimization (PSO): A stochastic algorithm inspired by the collective behavior of swarms, based on the movement of particles in the search space, combining their individual experience with the information shared by the group. It has demonstrated high efficiency in high-dimensional continuous spaces and exhibits good performance in optimizing machine learning models and neural networks (Kennedy & Eberhart, 1995). In this work, PSO was used in the group of models classified as Stochastic Continuous Search (SCS).
- Optuna: An optimization framework that combines a Bayesian approach with techniques such as Tree-structured Parzen Estimators (TPE), incorporating early pruning mechanisms that discard unpromising configurations at early stages, increasing computational efficiency without affecting the quality of the results. This makes it especially suitable for complex models or in resource-limited environments (Akiba, Sano, Yanase, Ohta, & Koyama, 2019). Like PSO, Optuna was applied to the group of models called Stochastic Continuous Search (SCS).

Although the optimization methods employed, such as Grid Search and Particle Swarm Optimization, as well as modern Bayesian optimization approaches implemented in frameworks such as Optuna, do not formally guarantee convergence to the global optimum, their use in this study follows widely accepted practices in the applied literature (Hutter, Kotthoff, & Vanschoren, 2019; Akiba, Sano, Yanase, Ohta, & Koyama, 2019). Grid Search provides exhaustive exploration in finite discrete spaces (Bergstra & Bengio, 2012), while PSO and Optuna have proven effective in optimizing complex models in continuous spaces, achieving significant improvements in predictive accuracy (Poli, Kennedy, & Blackwell, 2007; Zhang, Wang, & Ji, 2015; Anggraeni, et al., 2024). It should be emphasized that the objective of this work is not to compare optimization algorithms, but to evaluate the relative impact of different evaluation functions under the same optimizer, so that any limitation related to local optima affects all compared criteria equally (Kourentzes, Trapero, & Barrow, 2020).

### 3.3.6. Stage 6: HEF

Following the definition of the HEF, an algorithm was implemented to act as the evaluation function in the model optimization process. The function combines the $R^2$, MAE, and RMSE metrics through relative weights and incorporates progressive penalties based on established tolerance thresholds. Its computational complexity is $O(n + m)$, where n represents the number of training observations and m represents the number of generated predictions. Since it only involves constant additional operations, the function remains highly efficient and suitable for repeated application in optimization algorithms. Algorithm 1 receives the calculated metrics ($R^2$, MAE, and RMSE) as input, along with the model's predicted values (predictions) and the actual values from the training set (y_train).

**Algorithm 1.** Hierarchical Evaluation Function (HEF).

---

```
ALGORITHM HEF (predictions, r2, mae, rmse, y_train)

INPUT:
   predictions   → Predicted values from the model
   r2            → Coefficient of determination
   mae           → Mean Absolute Error
   rmse          → Root Mean Squared Error
   y_train       → Actual values from the training dataset

OUTPUT:
```

```
   score        → Penalized score representing the model's performance

BEGIN
   mean ← MEAN(y_train)
   IF mean ≈ 0 THEN
      mean ← small positive value (e.g., 1e-6)
   END IF

   mae_tolerance  ← RecommendMAETolerance(y_train)
   rmse_tolerance ← RecommendRMSETolerance(y_train)

   mae_threshold  ← mae_tolerance × mean
   rmse_threshold ← rmse_tolerance × mean

   weights    ← GetMetricWeights()
   base_score ← weights.r2 × (1 − r2)
              + weights.mae × (mae / mean)
              + weights.rmse × (rmse / mean)

   IF mae < mae_threshold AND rmse < rmse_threshold THEN
      score ← base_score
   ELSE IF mae < mae_threshold THEN
      score ← ApplyPenalty(base_score, 'penalize_level_1')
   ELSE IF rmse < rmse_threshold THEN
      score ← ApplyPenalty(base_score, 'penalize_level_2')
   ELSE
      score ← ApplyPenalty(base_score, 'penalize_level_3')
   END IF

   IF any value in predictions < 0 THEN
      score ← ApplyPenalty(base_score, 'penalize_level_4')
   END IF

   RETURN score
END
```

The RecommendMAETolerance(y) subroutine sets the tolerance threshold associated with the mean absolute error (MAE) from the training series represented by the variable *y*. The procedure calculates the coefficient of variation (CV), defined as the ratio between the standard deviation and the mean of the series, and assigns an adaptive tolerance level based on its value. When the CV is low (CV < 0.2), a strict threshold (0.1) is set to detect even small deviations; in series with moderate variability (0.2 ≤ CV < 0.5), the threshold is relaxed to 0.2; when dispersion is higher (0.5 ≤ CV < 1.0), it is extended to 0.3; and in highly volatile scenarios (CV ≥ 1.0), it is set to 0.4, allowing a more lenient evaluation (Bobko & Whybark, 1985). In this way, the subroutine dynamically adjusts the tolerance level according to the statistical characteristics of the series, thus reducing disproportionate penalties while balancing error control across stable and uncertain contexts.

**Algorithm 2.** RecommendMAETolerance.

```
SUBROUTINE RecommendMAETolerance(y)
   coeff_var ← STD(y) / MEAN(y)
   IF coeff_var < 0.2 THEN
      RETURN 0.1
   ELSE IF coeff_var < 0.5 THEN
      RETURN 0.2
   ELSE IF coeff_var < 1.0 THEN
```

```
    RETURN 0.3
  ELSE
    RETURN 0.4
  END IF
END SUBROUTINE
```

The RecommendRMSETolerance(y) subroutine defines the tolerance threshold associated with RMSE based on the coefficient of variation (CV) of the training series, calculated as the ratio between the standard deviation and the mean of the observed values. The CV adaptively determines the threshold: the lower the variability, the stricter the threshold, while in highly dispersed contexts, the tolerance range is widened. Since RMSE penalizes extreme errors more severely compared to MAE, the tolerance values defined for this subroutine are comparatively broader. Thus, in series with low variability ($CV < 0.2$), the threshold is set to 0.15, in moderate variability ($0.2 \leq CV < 0.5$) it is raised to 0.25, in high variability ($0.5 \leq CV < 1.0$) to 0.35, and in very volatile scenarios ($CV \geq 1.0$) to 0.4 (Bobko & Whybark, 1985). This design aims to strike a balance between detecting critical errors and maintaining robustness against fluctuations inherent in time series, thereby avoiding disproportionate penalties while ensuring an appropriate balance between sensitivity and stability.

**Algorithm 3.** RecommendRMSETolerance.

```
SUBROUTINE RecommendRMSETolerance(y)
  coeff_var ← STD(y) / MEAN(y)
  IF coeff_var < 0.2 THEN
    RETURN 0.15
  ELSE IF coeff_var < 0.5 THEN
    RETURN 0.25
  ELSE IF coeff_var < 1.0 THEN
    RETURN 0.35
  ELSE
    RETURN 0.4
  END IF
END SUBROUTINE
```

The ApplyPenalty(base_score, penalty_level) subroutine implements a progressive penalty scheme on the HEF base score, by applying increasing multipliers based on the degree of tolerance threshold violations: mild (×1.2) when a single metric exceeds the limit, moderate (×1.3) in case of multiple violations, severe (×1.5) for significant deviations, and extreme (×1.8) when negative or inconsistent predictions are detected. This mechanism ensures proportionality, avoiding excessive penalties for minor errors and assigning appropriate weight to critical violations. Complementarily, the selection of hierarchical weights for $R^2$, MAE, and RMSE is integrated into the same adaptive penalization approach, in line with Smith & Coit (Smith, Coit, Baeck, Fogel, & Michalewicz, 1997), who highlight the effectiveness of dynamically adjusting penalties according to the magnitude of the violation. Thus, assigning greater weight to MAE in low-variability scenarios or to RMSE in high-dispersion contexts enables the evaluation to be modulated according to robustness criteria. Furthermore, recent studies by Chang et al. (Chang, Li, & Lin, 2024) and Iacopini et al. (Iacopini, Ravazzolo, & Rossini, 2023) demonstrate that incorporating asymmetric penalties into loss functions and scoring rules reflects the unequal nature of forecasting errors, where not all errors have the same cost or impact on model validity. Overall, the combination of differentiated weights and progressive penalties reinforces the ability of HEF to integrate accuracy, robustness, and explainability, ensuring that model evaluation takes into account not only error magnitude, but also their relative importance and logical consistency.

**Algorithm 4.** ApplyPenalty.

```
SUBROUTINE ApplyPenalty(base_score, penalty_level)
   percentage ← MATCH penalty_level WITH:
      'penalize_level_1' → 1.2
      'penalize_level_2' → 1.3
      'penalize_level_3' → 1.5
      'penalize_level_4' → 1.8
   RETURN base_score × percentage
END SUBROUTINE
```

The GetMetricWeights() subroutine assigns differentiated weights to $R^2$ (1.0), MAE (1.0), and RMSE (0.5) to balance accuracy, robustness, and explanatory power within hierarchical model evaluation. Although no standard weighting has been formally proposed, the literature conceptually supports its use. Botchkarev (Botchkarev, 2019) argues that performance metrics serve distinct purposes and should be integrated differently into composite evaluation systems. In contrast, Hodson (Hodson, 2022) emphasizes that RMSE is more sensitive to extreme errors and therefore less robust to outliers, which justifies moderating its influence to improve predictive stability. Overall, this configuration enables HEF to prioritize explanatory power ($R^2$) and average accuracy (MAE), while accounting for the impact of extreme errors without allowing them to dominate the final evaluation.

**Algorithm 5.** GetMetricWeights.

```
SUBROUTINE GetMetricWeights()
   RETURN:
      r2_weight   ← 1.0
      mae_weight  ← 1.0
      rmse_weight ← 0.5
END SUBROUTINE
```

The assignment of hierarchical weights to the selected metrics follows a conservative methodological criterion aimed at balancing explanatory power, average accuracy, and sensitivity to extreme errors (Makridakis, Spiliotis, & Assimakopoulos, 2018; Hyndman & Koehler, 2006). Specifically, equal weights are assigned to $R^2$ and MAE, prioritizing variance explanation and average error reduction, while the weight for RMSE is moderated due to its greater sensitivity to outliers (Chai & Draxler, 2014; Willmott, Matsuura, & Robeson, 2009). Previous studies have shown that the weighted integration of metrics with different properties enables the construction of more stable and robust evaluation criteria than those based on a single indicator (Hodson, 2022; Vogt, Remmen, Lauster, Fuchs, & Müller, 2018). In this study, the weights are kept fixed across all experiments to isolate the structural effect of the hierarchical function and avoid introducing additional variability from configuration-specific tuning. The exploration of alternative or adaptive weighting schemes is explicitly identified as a direction for future research.

### 3.3.7. Stage 7: MAE Evaluation Function

The implementation of the evaluation function based exclusively on the mean absolute error, called MAEF, is a classical evaluation function widely used in the literature for parameter and hyperparameter optimization, whose main purpose is to serve as a reference against the HEF function. Algorithm 6 receives the calculated MAE value and returns it directly without applying additional modifications, thus ensuring full compatibility with other evaluation schemes. Limited to a simple assignment and return, its computational complexity is $O_{(1)}$, making it a highly efficient implementation computationally.

**Algorithm 6.** MAE Evaluation Function (MAEF).

```
ALGORITHM MAEF(mae)
INPUT:
   mae        → Mean Absolute Error
OUTPUT:
   best_value → Value selected as the base evaluation metric
BEGIN
   best_value ← mae
   RETURN best_value
END
```

### 3.3.8. Stage 8: Results

At this stage, the results of the experimental runs are obtained, including predictions generated during training, as well as traditional evaluation metrics ($R^2$, MAE, and RMSE). Based on the data, where $y_t$ corresponds to the actual value at time t, $\hat{y}_t$ to the value predicted by the model, n to the number of observations, and h to the prediction horizon, complementary metrics such as Global Relative Accuracy (GRA), RMSSE, and MASE are also calculated.

Global Relative Accuracy (GRA): In the context of demand forecasting, point-in-time metrics such as MAE and RMSE enable the evaluation of performance period by period by comparing observed and predicted values. However, when the objective is to measure the cumulative accuracy of the model in terms of total projected volume versus actual volume, these metrics may not adequately capture systematic deviations over the evaluation horizon. To this end, we propose Global Relative Accuracy (Equation 8), which measures the relative difference between the cumulative totals of forecasts and observations, using absolute values to avoid numerical cancellations in series that include negative values. The result is interpreted as an overall accuracy index, where values close to 1 reflect a strong correspondence between the total forecast and observed volumes.

$$\text{GRA} = 1 - \frac{\left|\sum_{t=1}^{n}|\hat{y}_t| - \sum_{t=1}^{n}|y_t|\right|}{\sum_{t=1}^{n}|y_t|} \tag{8}$$

The design of this indicator is inspired by the approach proposed by (Ha, Seok, & Ok, 2018), who introduced the Cumulative Absolute Forecast Error (CAFE) metric to capture total accumulated error in aggregate planning scenarios. CAFE evaluates the sum of absolute errors per period and allows for the inclusion of penalties related to costs derived from overestimation or underestimation. In contrast, GRA focuses on the alignment between forecasted and observed totals, providing a synthetic view of overall model performance. This metric is also consistent with the recommendations of Adhikari et al. (2017), who emphasize the importance of minimizing cumulative deviation from actual sales to achieve a more robust evaluation of forecasting model performance.

RMSSE: The Scaled Root Mean Square Error (Equation 9) uses the error of a naive model as a reference to scale the errors of the evaluated model. Like RMSE, RMSSE penalizes large errors more severely, allowing for an appropriate comparison between different series while maintaining sensitivity to significant errors.

$$RMSSE = \sqrt{\frac{\frac{1}{h}\sum_{t=1}^{h}(y_t - \hat{y}_t)^2}{\frac{1}{n-1}\sum_{t=2}^{n}(y_t - y_{t-1})^2}} \tag{9}$$

MASE: The Mean Absolute Scaled Error (Equation 10) is a robust measure that normalizes MAE with respect to the mean absolute error of a reference model, typically a naive forecast based on the last observed value. This formulation facilitates interpretation: values below 1 indicate that the model outperforms the naive forecast, while values above 1 reflect poorer performance. Unlike percentage-based measures, MASE is not affected by values close to zero in the series, making it a reliable indicator in contexts with high variability.

$$MASE = \frac{\frac{1}{n}\sum_{t=1}^{n}|y_t - \hat{y}_t|}{\frac{1}{n-1}\sum_{t=2}^{n}|y_t - y_{t-1}|} \tag{10}$$

It should be emphasized that the penalty thresholds and hierarchical weights adopted in HEF were derived directly from the literature and kept fixed throughout all experiments. Their inclusion in this study served the purpose of ensuring methodological consistency and comparability, rather than exploring sensitivity to alternative configurations. Consequently, no illustrative examples or additional experiments are presented at this stage, since such analyses are explicitly considered as part of future work (see Section 6.4).

#### 3.3.9. Experimental Protocol

For the empirical evaluation, an execution protocol was applied. Executed procedures are marked with an "X" and those not executed with a hyphen "–". The ID column corresponds to the execution code associated with the dataset, while column 1 identifies the selected dataset (Walmart, M5, M4, or M3). Column 2 indicates the training and testing configurations considered (91:9, 80:20, and 70:30), and column 3 specifies the participating models defined in Stage 3. Column 4 reflects the execution of the models under fixed configurations, as outlined in Stage 4. Column 5 is divided into three subcolumns: 5(a) represents optimization using Grid Search, 5(b) optimization using Particle Swarm Optimization (PSO), and 5(c) optimization using Optuna. Column 6 records the application of the HEF function along with the corresponding optimizer, while column 7 records the MAEF function. Column 8 contains the indicators obtained in each run, and column 9 shows the number of repetitions performed for each time series. In this study, each series was run 21 times to ensure statistical stability, reduce random variability, and guarantee representativeness. The complete execution scheme is presented in Appendix B (Table B1).

For each set of results associated with the time series, the normality of the distributions was verified, and based on this, the most appropriate statistical test was selected, using either parametric or nonparametric methods as applicable. This procedure ensured the consistency and robustness of the observed differences, thereby preventing the findings from being dependent on a single run or random fluctuations. Each difference detected and statistically validated was defined as a case, understood as a significant change in the result.

Once the iterations were completed, an analysis was performed by dataset and by individual time series, considering each series as an independent case. The objective was to identify differences in performance metrics derived from the combination of evaluation functions and optimization methods, with a particular focus on the impact of the HEF function. Statistical comparisons were used to assess the significance of the observed changes, allowing the determination of whether the incorporation of a custom evaluation function generates systematic improvements, whether these improvements vary by model type, or whether they are associated with specific patterns in each dataset.

#### 3.3.10. Hardware and Software

The computational implementation of the forecasting models was carried out using a wide variety of scientific and machine learning libraries in Python. Statistical and time series models, such as ARIMA and SES, were implemented through the Statsmodels library (Seabold, Perktold, & others, 2010), while the LSR, RR, ENR, HR, RPL, KNN, SVR, RFR, and GBR models, as well as Bayesian regressors, were developed using Scikit-learn (Pedregosa, et al., 2011). The XGBoost model was implemented with the XGBoost library (Chen & Guestrin, 2016), and CatBoost was implemented using its official API (Prokhorenkova, Gusev, Vorobev, Dorogush, & Gulin, 2018). On the other hand, neural network models, including MLP and LSTM, were built using TensorFlow and its high-level interface Keras (Chollet, 2015; Abadi, et al., 2016). For data preparation, manipulation, and analysis, Pandas (Team, 2020), NumPy (Harris, et al., 2020), and SciPy (Virtanen, et al., 2020) were employed. Result visualization was performed using Matplotlib (Hunter, 2007) and Seaborn (Waskom, 2021). Hyperparameter optimization was addressed through search algorithms such as PSO and Optuna (Akiba, Sano, Yanase, Ohta, & Koyama, 2019).

Full details of the experimental environment, optimizer configurations, and parameter ranges explored are reported in Appendix C (Table C1).

## 4. Results

By applying the described protocol, results were obtained for each combination of datasets, training and testing configurations, models, evaluation functions, and optimization methods. The following section presents comparative results evaluating the impact of HEF relative to traditional approaches, demonstrating its consistent ability to improve predictions. Reported values correspond to the total number of statistically validated cases identified for the Walmart, M3, M4, and M5 datasets.

Difference-of-proportions tests yielded highly significant Z values ($p \approx 0.0$) across all datasets and optimizers, confirming that the observed improvements are attributable to the evaluation function rather than the search method. This consolidated evidence demonstrates that HEF consistently outperforms MAEF on global metrics, including $R^2$, GRA, and RMSSE.

It should be noted that the values reported in Tables 4, 5, and 6 represent the number of statistically validated cases in which one evaluation function outperformed the other, rather than absolute metric values.

### *4.1. Training 91% and Testing 9%*

The evaluation with a 91:9 partition was conducted using Grid Search, PSO, and Optuna. Detailed numerical comparisons are presented in Appendix D (Tables D1–D3). A consolidated view is presented in Table 4, highlighting the comparative outcomes across the various optimization methods.

**Table 4.** Summary of comparative results across optimization methods for the 91:9 partition.

| Metrics | Grid Search | | PSO | | Optuna | |
|---|---|---|---|---|---|---|
| | Improves HEF | Improves MAEF | Improves HEF | Improves MAEF | Improves HEF | Improves MAEF |
| $R^2$ | **1673** (↑) | 0 | **1639** (↑) | 288 | **2518** (↑) | 274 |
| MAE | 0 | **1386** (↑) | 270 | **1017** (↑) | 248 | **1685** (↑) |
| RMSE | 0 | **514** (↑) | **969** (↑) | 235 | **1398** (↑) | 360 |
| RMSSE | 0 | **424** (↑) | **1230** (↑) | 272 | **1766** (↑) | 324 |
| MASE | 0 | **1458** (↑) | 238 | **1073** (↑) | 279 | **1799** (↑) |
| GRA | **1281** (↑) | 0 | **1207** (↑) | 429 | **1869** (↑) | 538 |
| Execution Time | 386 | **4574** (↑) | 480 | **2723** (↑) | 1502 | **2464** (↑) |

The results confirm that HEF consistently improves $R^2$, GRA, RMSE, and RMSSE compared to MAEF, regardless of the optimizer applied, while also demonstrating effi-ciency in execution time in several scenarios.

To illustrate these outcomes, Figures 3, 4, and 5 (inserted here) present violin plots for Grid Search, PSO, and Optuna, respectively, showing the percentage distribution of improvements by metric. Positive values indicate advantages for HEF, while negative values reflect benefits for MAEF.

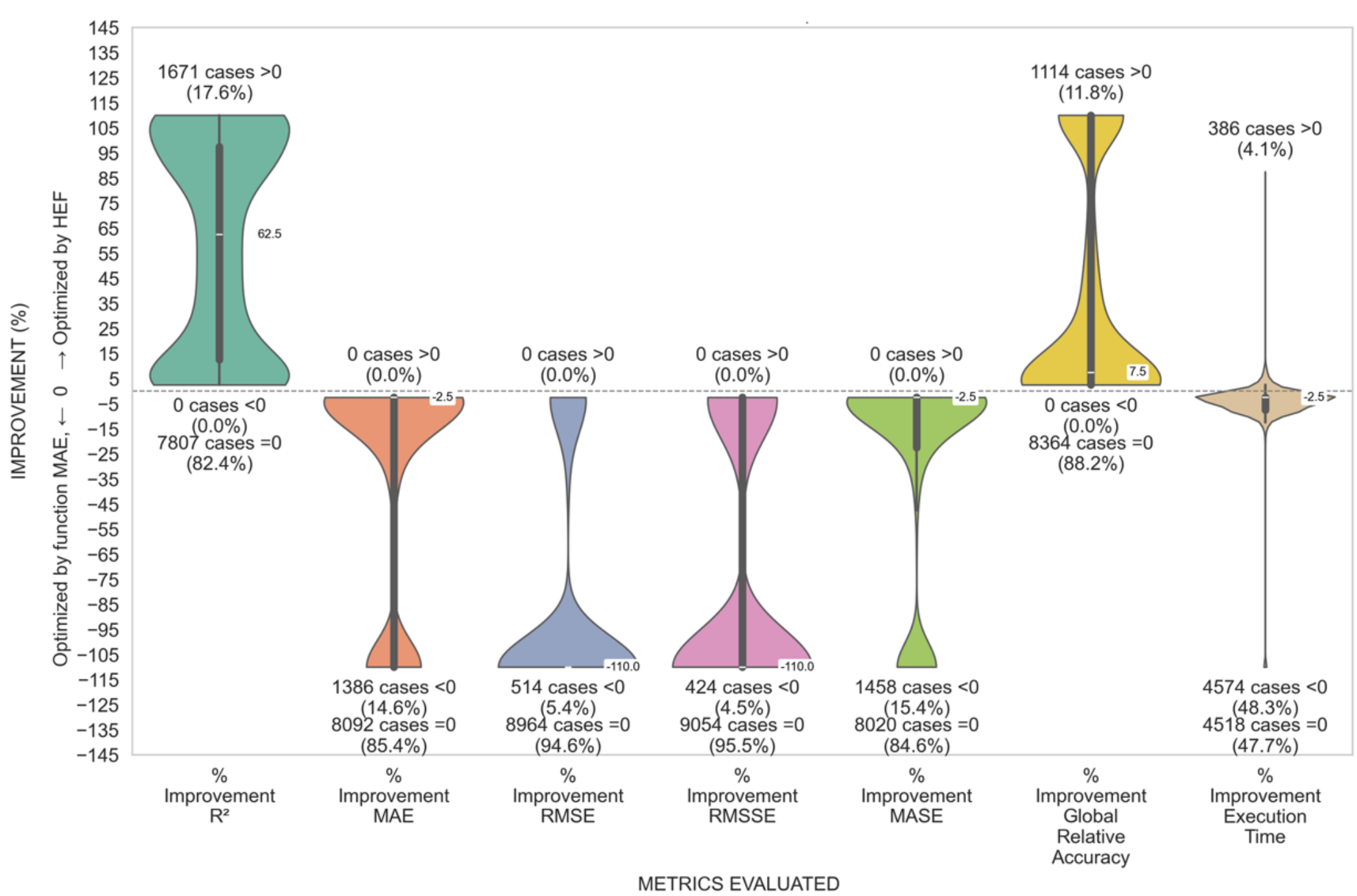

**Figure 3.** Distribution of percentage improvement by metric using evaluation function optimized by MAEF v/s optimized by HEF configuration training and testing 91:9 | optimizer used Grid Search.

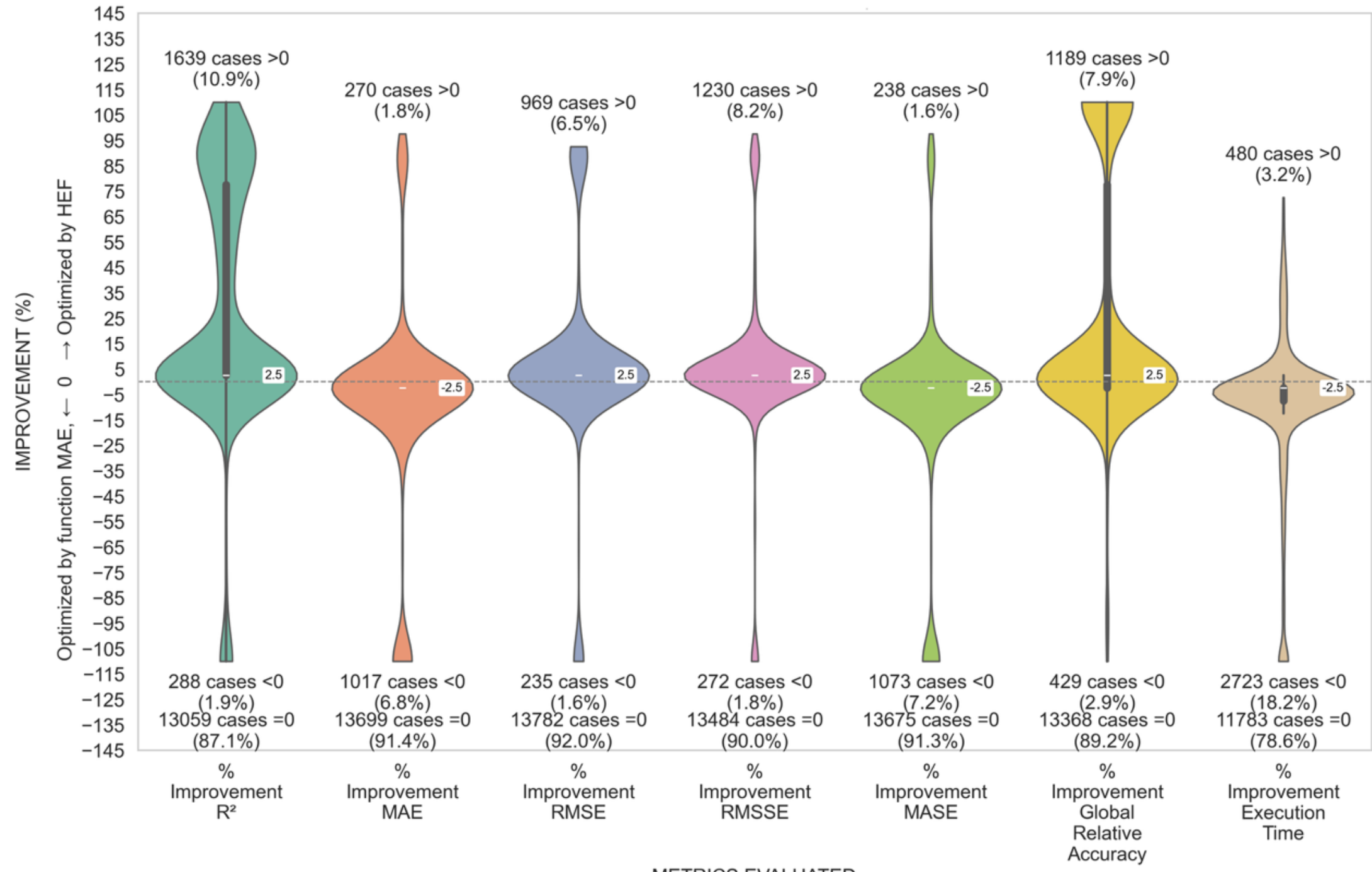

**Figure 4.** Distribution of percentage improvement by metric using evaluation function optimized by MAEF v/s optimized by HEF configuration training and testing 91:9 | optimizer used PSO.

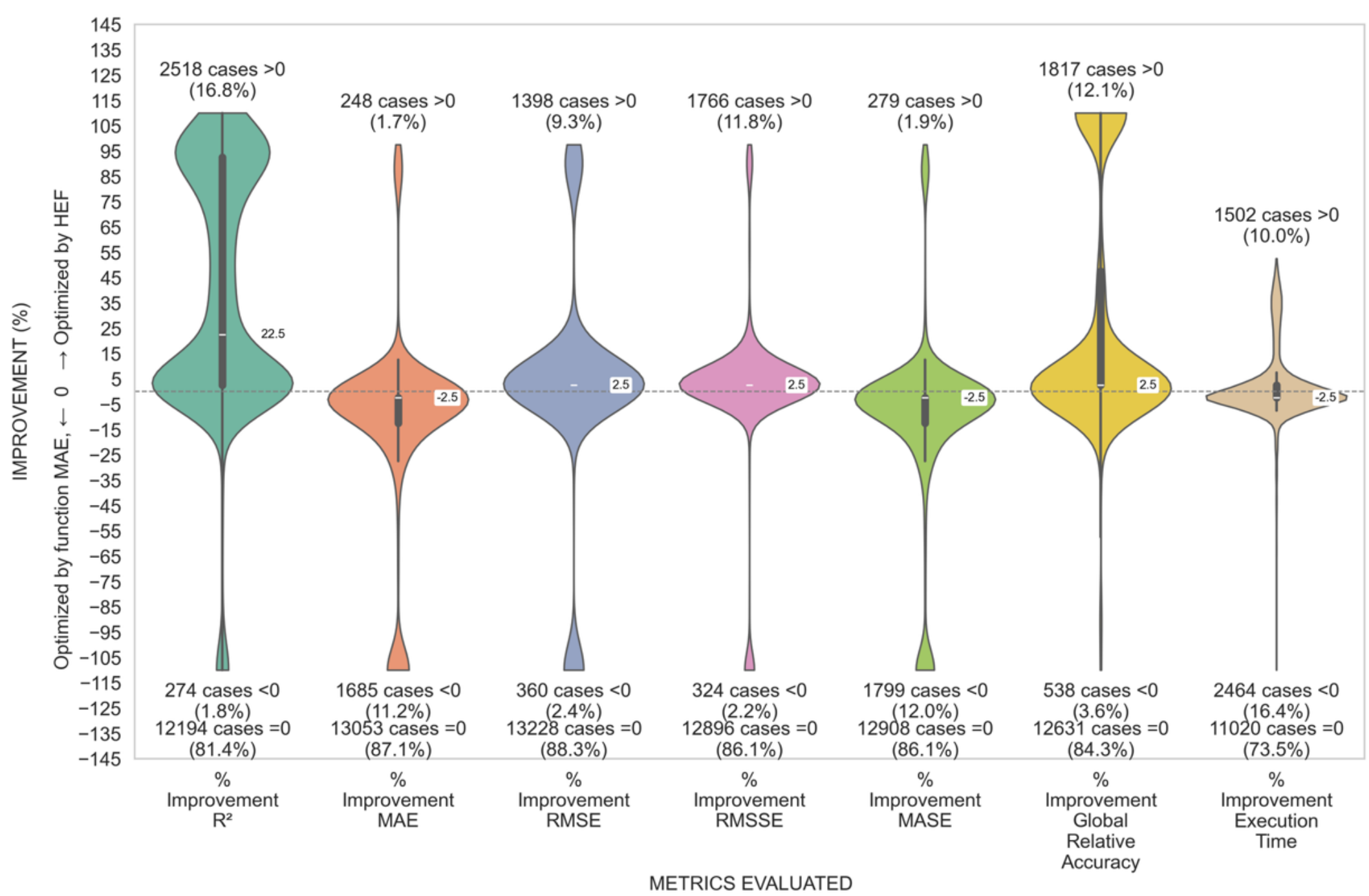

**Figure 5.** Distribution of percentage improvement by metric using evaluation function optimized by MAEF v/s optimized by HEF configuration training and testing 91:9 | optimizer used Optuna.

The consolidated results correspond to the four instances: Walmart, M3, M4, and M5. The disaggregated results for each dataset are provided in Appendix D (Tables D4–D7).

In all cases, the difference-of-proportions tests yielded highly significant values ($p \approx 0.0$), confirming that the observed improvements originated from the evaluation function rather than from the optimization method. This global evidence rules out the influence of chance and consolidates the superiority of HEF over MAEF in terms of explanatory capacity, overall accuracy, and robustness against extreme errors. The disaggregated results by dataset are presented in Appendix D (Table D8), which reports the Z and p values obtained from the difference-of-proportions tests applied to significant cases derived from the metrics $R^2$, GRA, RMSE, and RMSSE, in the comparison between HEF and MAEF.

Furthermore, Figure 6 illustrates case-level comparisons between models optimized with MAEF and those optimized with HEF using Optuna, exposing the behavior of HEF across individual models.

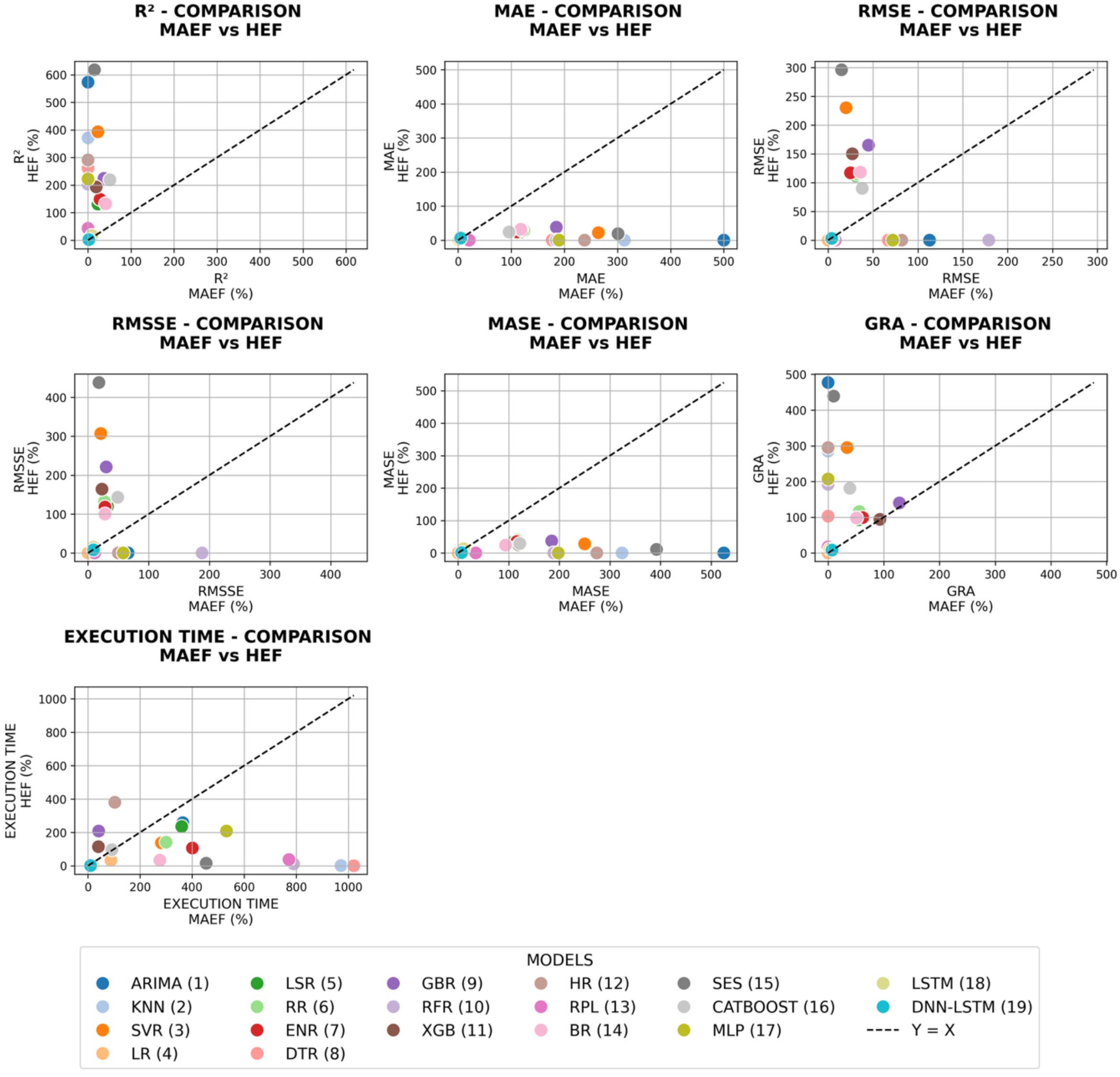

**Figure 6.** Distribution of percentage improvement by metric and model using evaluation function optimized by MAEF v/s optimized by HEF configuration training and testing 91:9 | optimizer used Optuna.

## *4.2. Training 80% and Testing 20%*

The evaluation with an 80:20 partition was conducted using Grid Search, PSO, and Optuna. The detailed results are provided in Appendix E (Tables E1–E3). A consolidated summary is presented in Table 5, highlighting the comparative outcomes across the various optimization methods.

**Table 5.** Summary of comparative results across optimization methods for the 80:20 partition.

| Metrics | Grid Search | | PSO | | Optuna | |
|---|---|---|---|---|---|---|
| | Improves HEF | Improves MAEF | Improves HEF | Improves MAEF | Improves HEF | Improves MAEF |
| $R^2$ | **1898** (↑) | 0 | **1794** (↑) | 221 | **2718** (↑) | 289 |
| MAE | 0 | **1670** (↑) | 247 | **1163** (↑) | 254 | **1862** (↑) |
| RMSE | 0 | **624** (↑) | **1205** (↑) | 230 | **1574** (↑) | 361 |
| RMSSE | 0 | **537** (↑) | **1394** (↑) | 237 | **1894** (↑) | 370 |
| MASE | 0 | **1733** (↑) | 248 | **1194** (↑) | 275 | **1931** (↑) |
| GRA | **1448** (↑) | 0 | **1148** (↑) | 634 | **1692** (↑) | 914 |
| Execution Time | 523 | **4040** (↑) | 549 | **2364** (↑) | 1264 | **2599** (↑) |

The findings confirm that HEF consistently improves R², GRA, RMSE, and RMSSE compared to MAEF, independent of the optimizer. Although execution times tend to increase with PSO and Optuna, the performance gains achieved by HEF justify this computational cost.

The distributions of improvements across metrics are visualized in Figures 7, 8, and 9 for Grid Search, PSO, and Optuna, respectively.

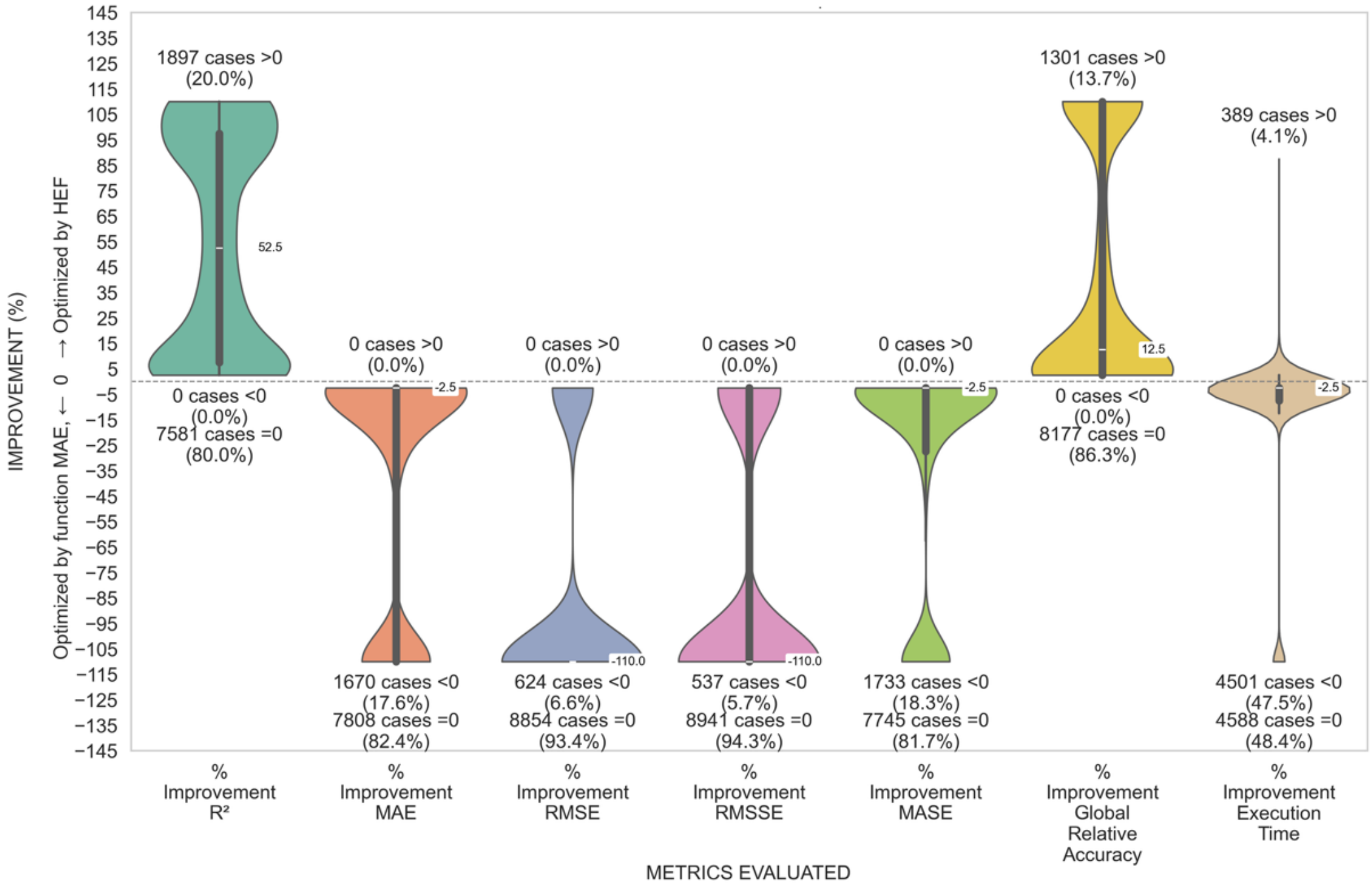

**Figure 7.** Distribution of percentage improvement by metric using evaluation function optimized by MAEF v/s optimized by HEF configuration training and testing 80:20 | optimizer used Grid Search.

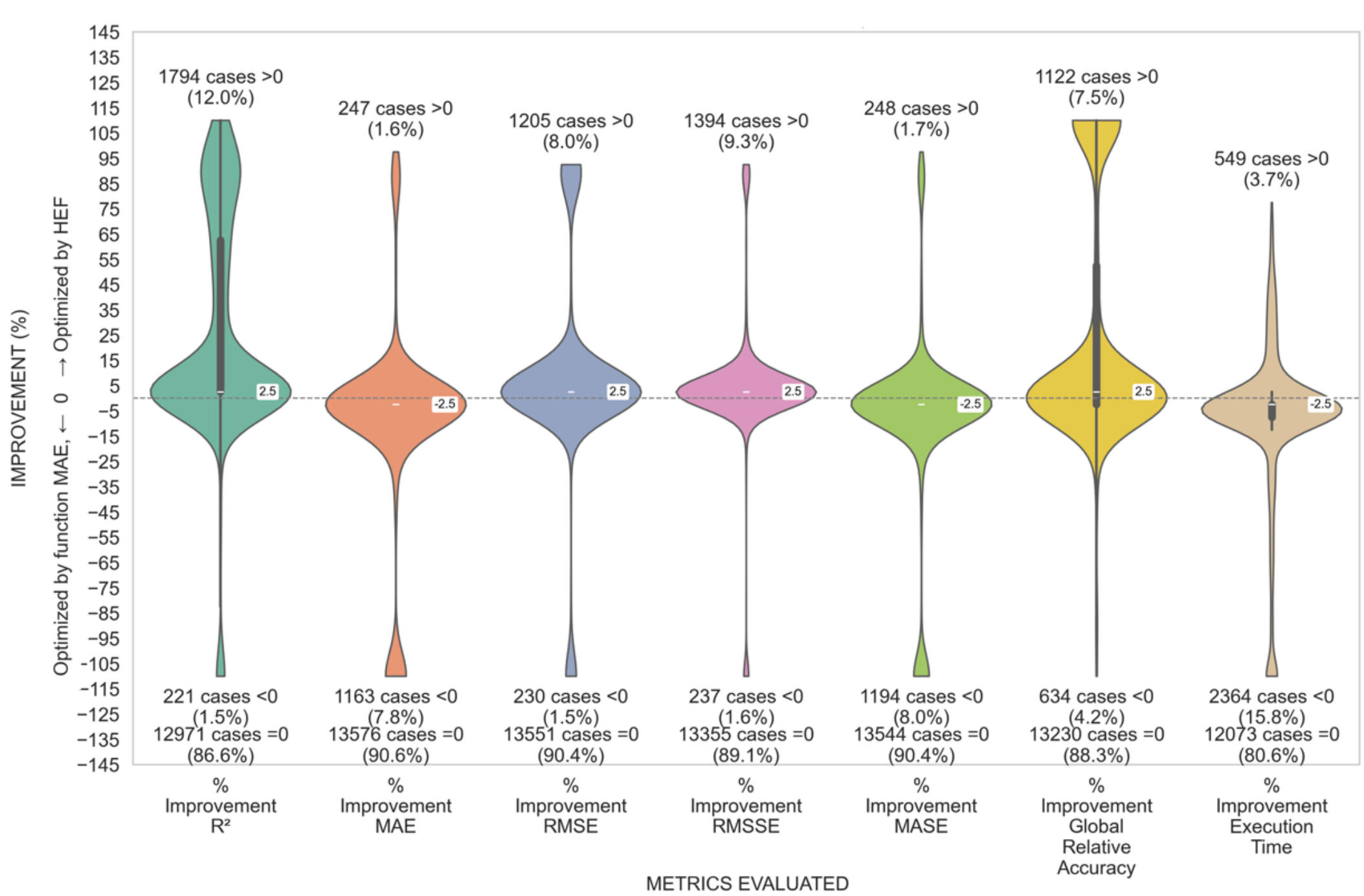

**Figure 8.** Distribution of percentage improvement by metric using evaluation function optimized by MAEF v/s optimized by HEF configuration training and testing 80:20 | optimizer used PSO.

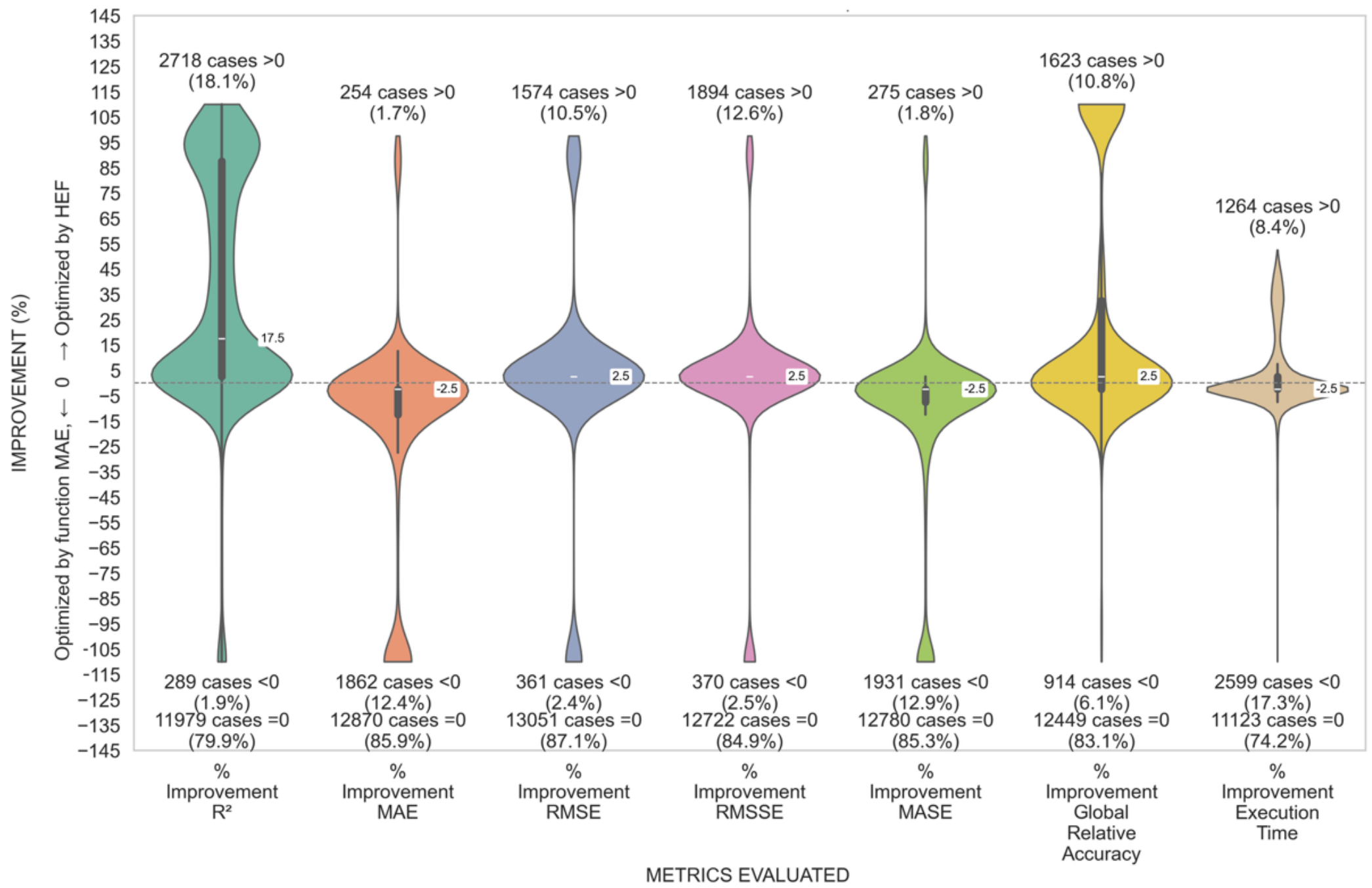

**Figure 9.** Distribution of percentage improvement by metric using evaluation function optimized by MAEF v/s optimized by HEF configuration training and testing 80:20 | optimizer used Optuna.

The consolidated results correspond to the four instances: Walmart, M3, M4, and M5. The disaggregated results for each dataset are provided in Appendix E (Tables E4–E7).

In all cases, the difference-of-proportions tests yielded highly significant values ($p \approx 0.0$), confirming that the observed improvements originated from the evaluation function rather than from the optimization method. This global evidence rules out the influence of chance and consolidates the superiority of HEF over MAEF in terms of explanatory capacity, overall accuracy, and robustness against extreme errors. The disaggregated results by dataset are presented in Appendix E (Table E8), which reports the Z and p values obtained from the difference-of-proportions tests applied to significant cases derived from the metrics $R^2$, GRA, RMSE, and RMSSE, in the comparison between HEF and MAEF.

Furthermore, Figure 10 illustrates case-level comparisons between models optimized with MAEF and those optimized with HEF using Optuna, exposing the behavior of HEF across individual models.

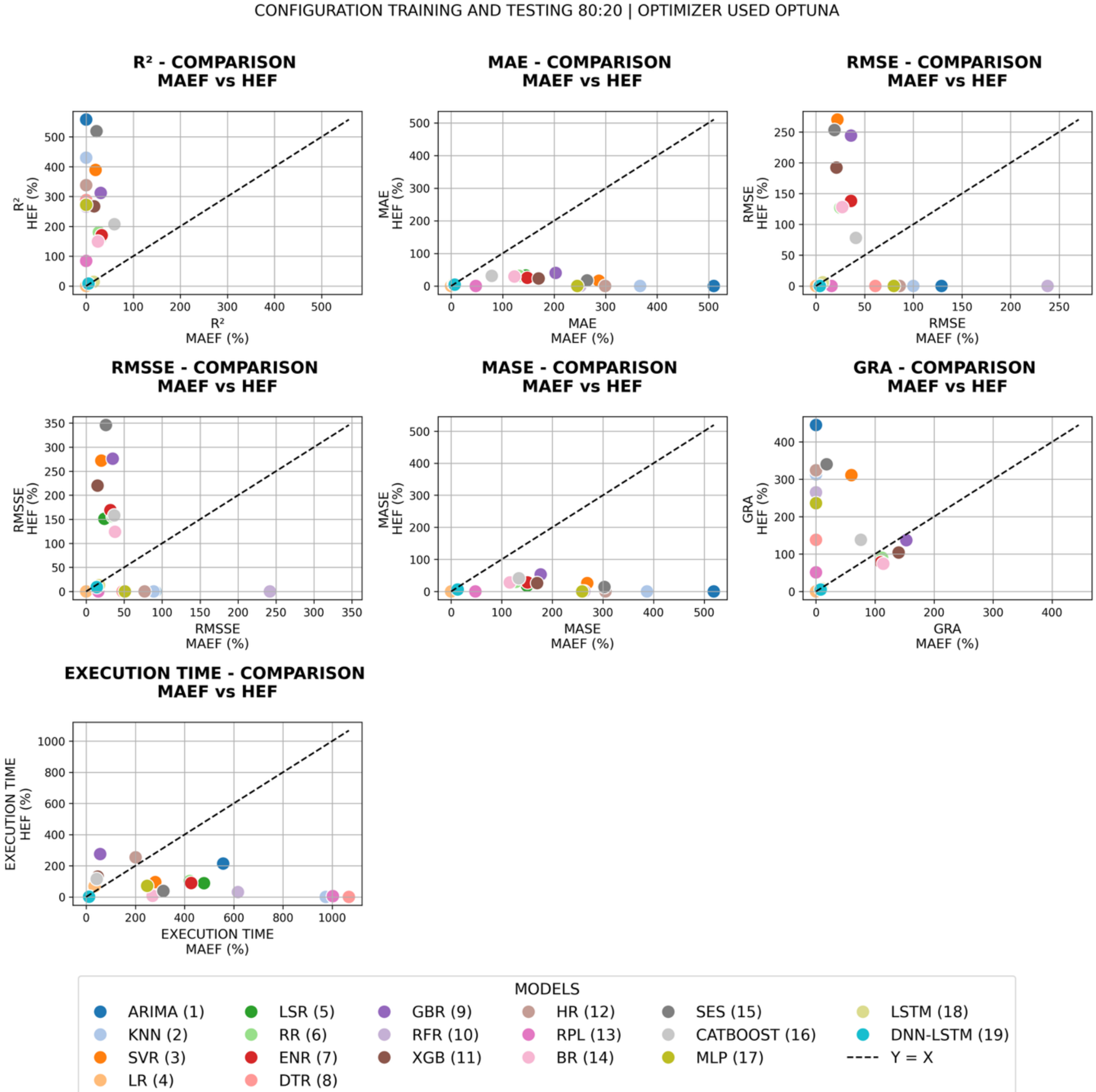

**Figure 10.** Distribution of percentage improvement by metric and model using evaluation function optimized by MAEF v/s optimized by HEF configuration training and testing 80:20 | optimizer used Optuna.

## 4.3. Training 70% and Testing 30%

The evaluation with a 70:30 partition was conducted using Grid Search, PSO, and Optuna. Detailed numerical comparisons are provided in Appendix F (Tables F1–F3). A consolidated summary is presented in Table 6, highlighting the comparative outcomes across the various optimization methods.

**Table 6.** Summary of comparative results across optimization methods for the 70:30 partition.

| Metrics | Grid Search | | PSO | | Optuna | |
|---|---|---|---|---|---|---|
| | Improves HEF | Improves MAEF | Improves HEF | Improves MAEF | Improves HEF | Improves MAEF |
| $R^2$ | **2035** (↑) | 0 | **1840** (↑) | 248 | **2709** (↑) | 254 |
| MAE | 0 | **1744** (↑) | 235 | **1152** (↑) | 249 | **1871** (↑) |
| RMSE | 0 | **651** (↑) | **1145** (↑) | 240 | **1553** (↑) | 345 |
| RMSSE | 0 | **574** (↑) | **1355** (↑) | 249 | **1923** (↑) | 309 |
| MASE | 0 | **1838** (↑) | 242 | **1151** (↑) | 253 | **1827** (↑) |
| GRA | **1512** (↑) | 0 | **1140** (↑) | 621 | **1725** (↑) | 878 |
| Execution Time | 1314 | **3384** (↑) | 599 | **2074** (↑) | 1175 | **2945** (↑) |

The results demonstrate that HEF consistently improves $R^2$, GRA, RMSE, and RMSSE relative to MAEF, regardless of the optimizer used. Execution times were higher in several cases, but the robustness of HEF justifies this additional computational cost.

The distributions of improvements are shown in Figures 11, 12, and 13 for Grid Search, PSO, and Optuna, respectively, while Figure 14 provides a case-level comparison under Optuna. Positive values indicate advantages for HEF, while negative values reflect advantages for MAEF.

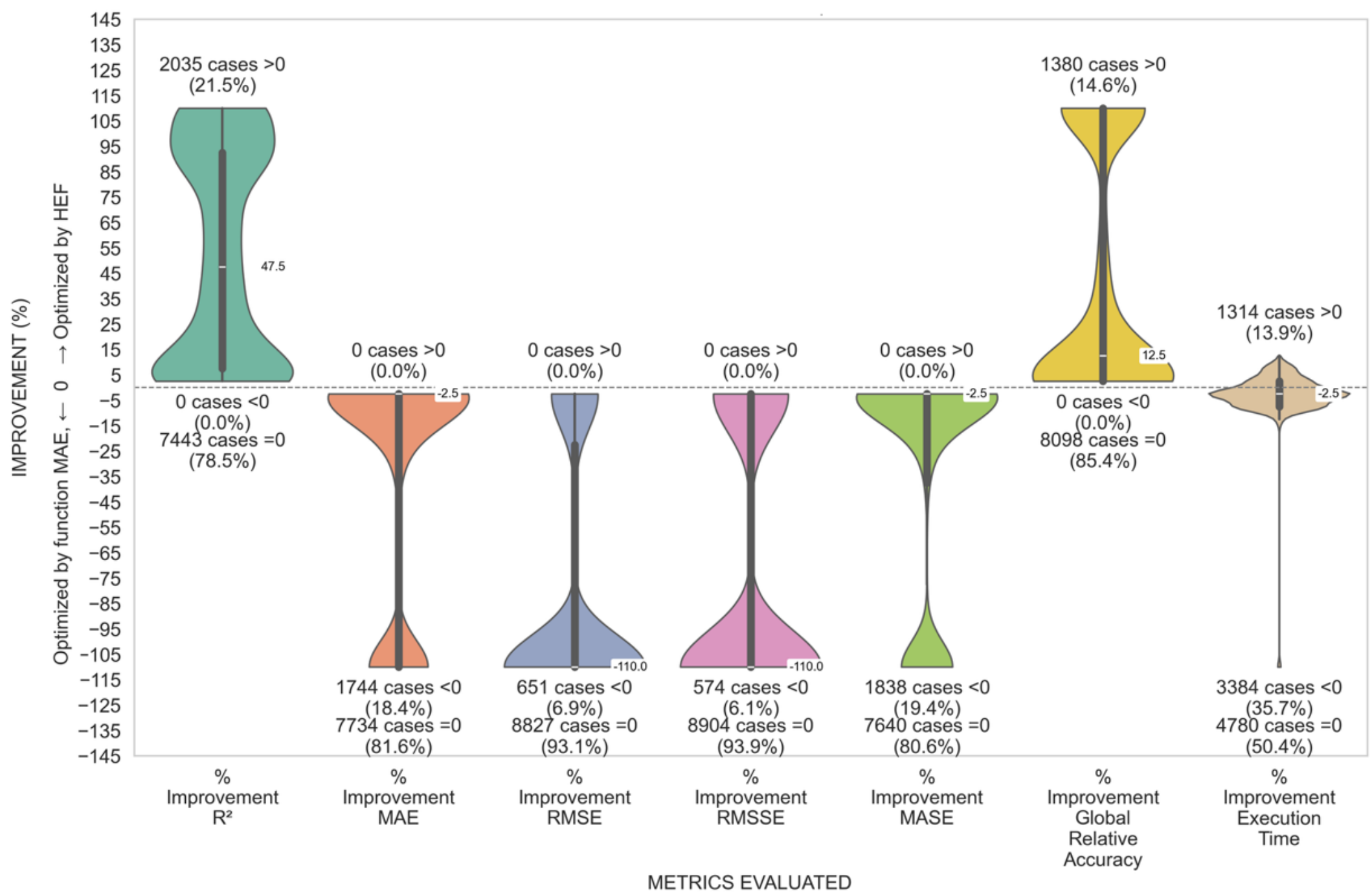

**Figure 11.** Distribution of percentage improvement by metric using evaluation function optimized by MAEF v/s optimized by HEF configuration training and testing 70:30 | optimizer used Grid Search.

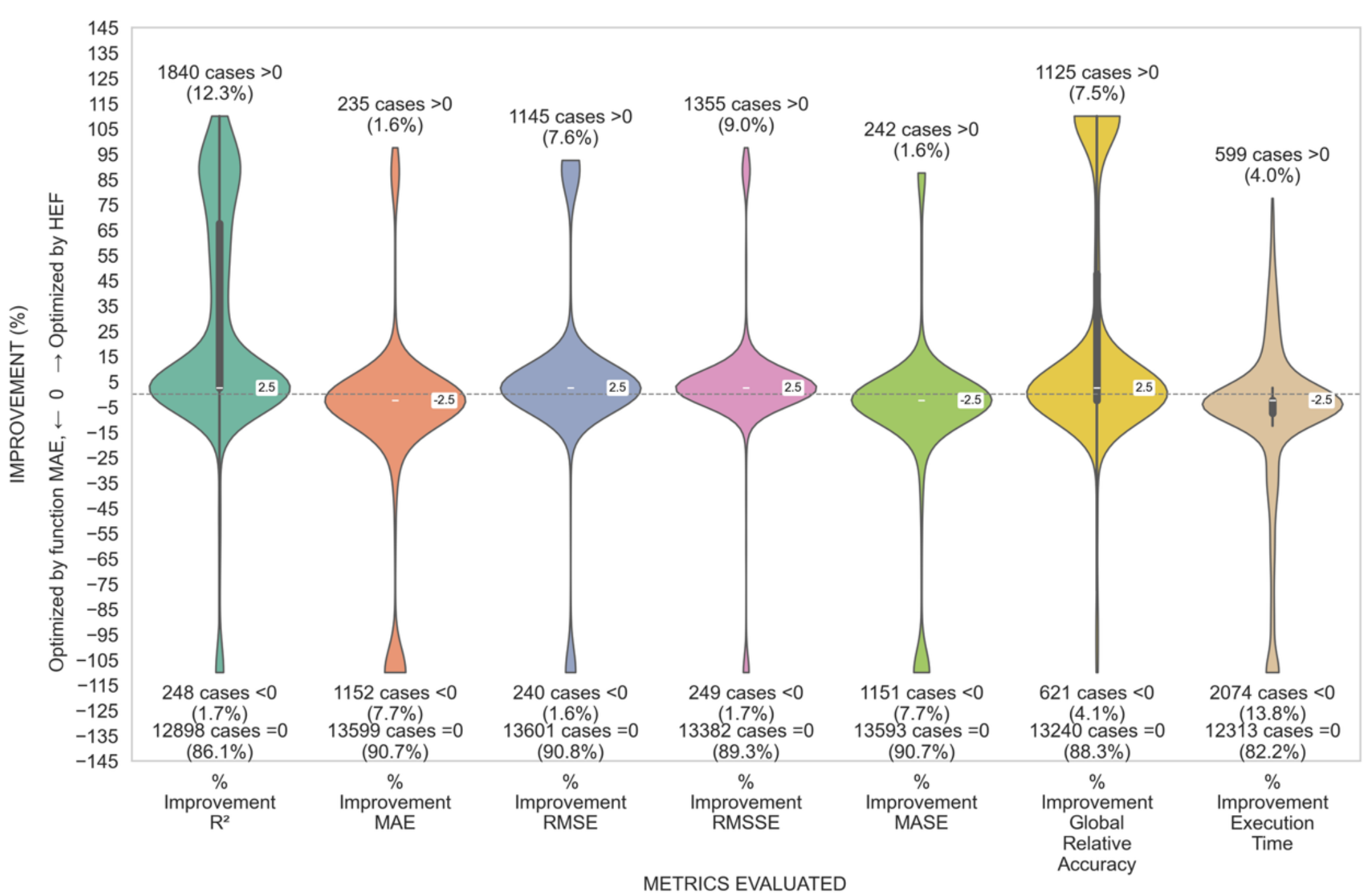

**Figure 12.** Distribution of percentage improvement by metric using evaluation function optimized by MAEF v/s optimized by HEF configuration training and testing 70:30 | optimizer used PSO.

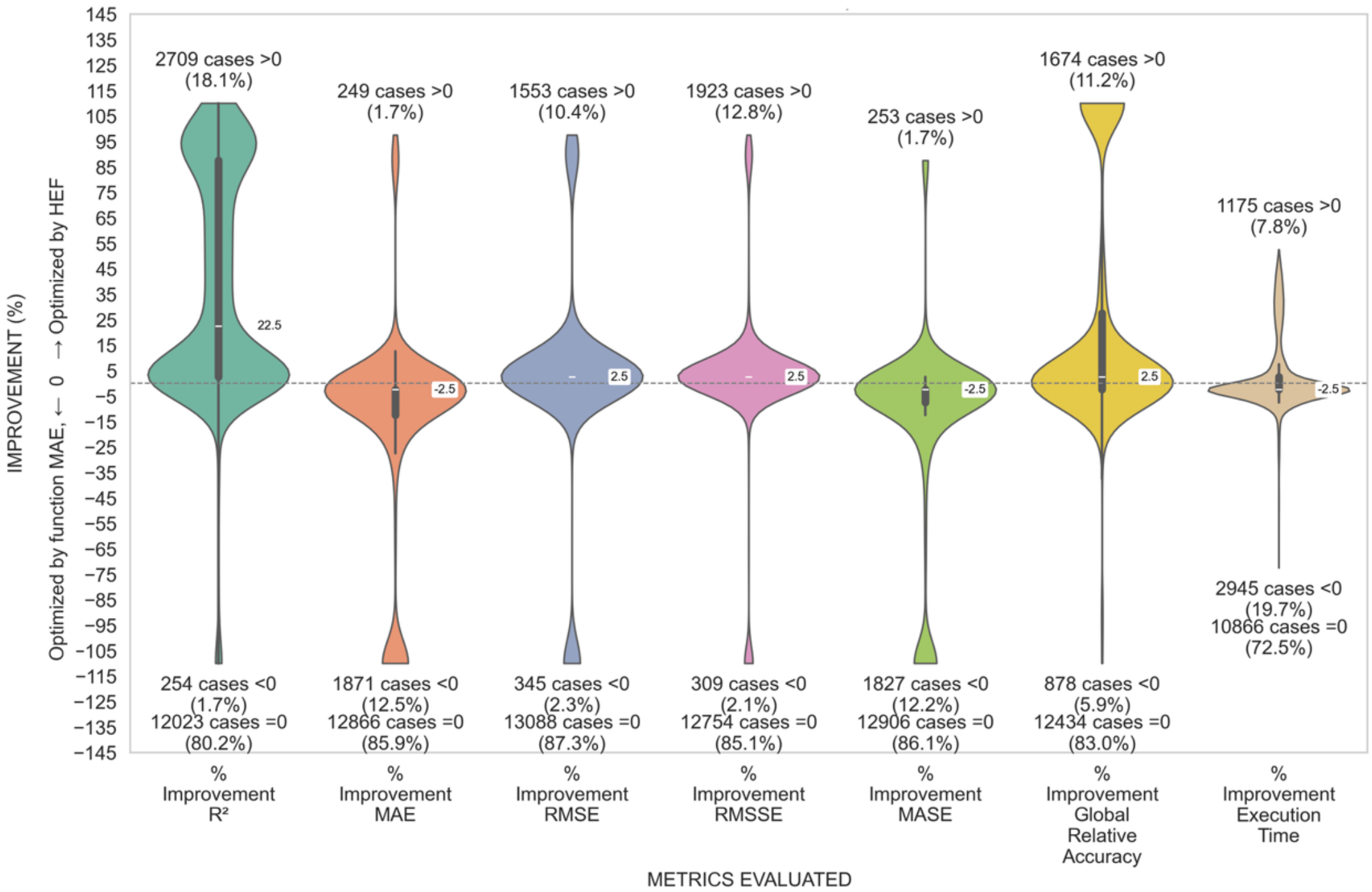

**Figure 13.** Distribution of percentage improvement by metric using evaluation function optimized by MAEF v/s optimized by HEF configuration training and testing 70:30 | optimizer used Optuna.

The disaggregated results for each dataset are provided in Appendix F (Tables F4–F7), ensuring reproducibility and completeness of the evaluation.

In all cases, the difference-of-proportions tests yielded highly significant values ($p \approx 0.0$), confirming that the observed improvements originated from the evaluation function rather than from the optimization method. This global evidence rules out the influence of chance and consolidates the superiority of HEF over MAEF in terms of explanatory capacity, overall accuracy, and robustness against extreme errors. The disaggregated results by dataset are presented in Appendix F (Table F8), which reports the Z and p values obtained from the difference-of-proportions tests applied to significant cases derived from the metrics $R^2$, GRA, RMSE, and RMSSE, in the comparison between HEF and MAEF.

Furthermore, Figure 14 illustrates case-level comparisons between models optimized with MAEF and those optimized with HEF using Optuna, exposing the behavior of HEF across individual models.

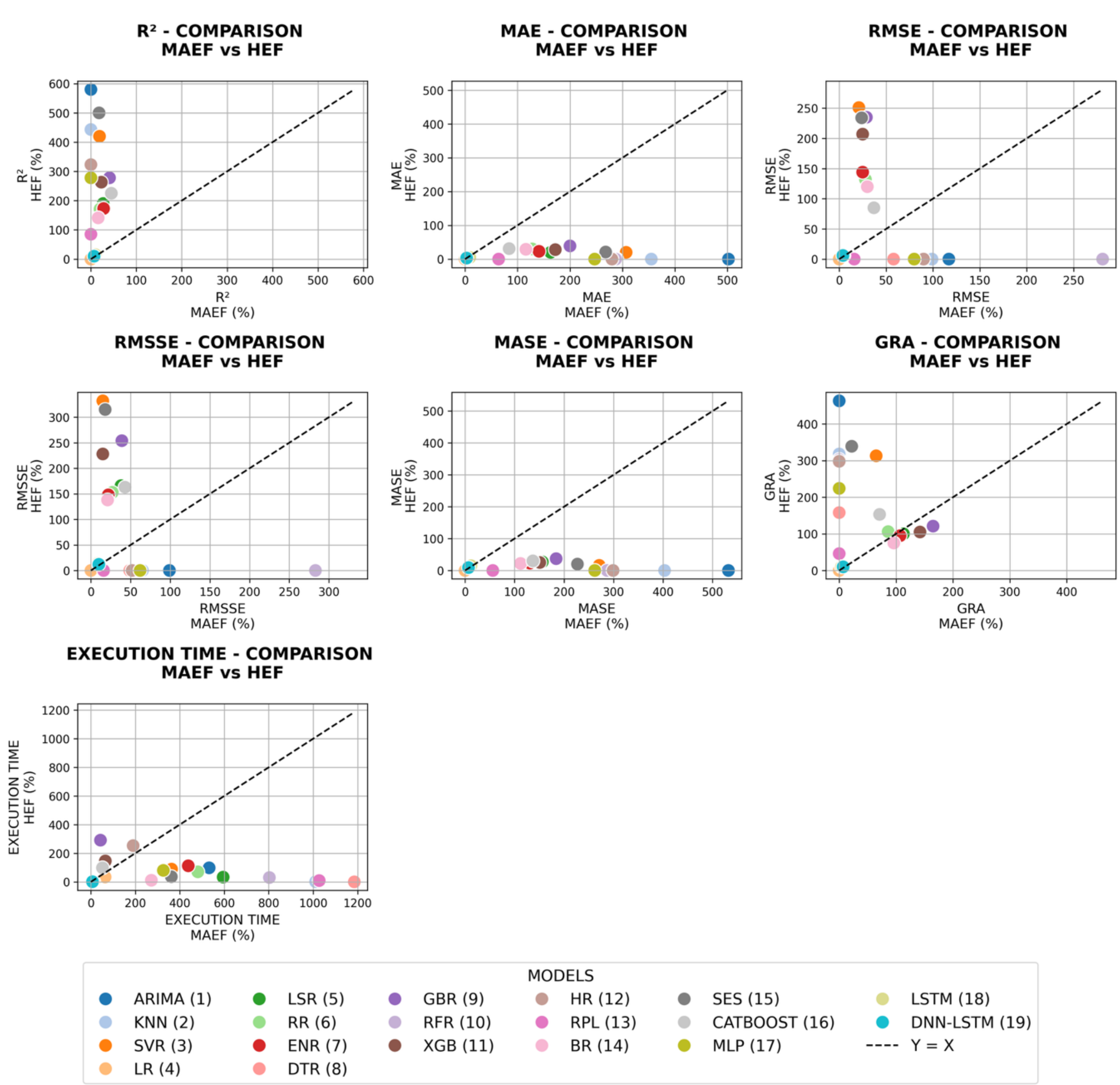

**Figure 14.** Distribution of percentage improvement by metric and model using evaluation function optimized by MAEF v/s optimized by HEF configuration training and testing 70:30 | optimizer used Optuna.

## 5. Discussion

The results obtained clearly isolate the effect of the evaluation function on the performance of demand prediction models. Furthermore, the consistency of the results observed across different optimizers, Grid Search, PSO, and Optuna, indicates that the observed improvements are primarily attributable to the evaluation function rather than to the search method itself. This behavior reinforces the robustness of the proposed approach and aligns with previous studies highlighting the importance of the evaluation criterion in determining final model quality, regardless of the underlying optimization algorithm (Khan & Osińska, 2023; Papacharalampous, Tyralis, & Koutsoyiannis, 2018; Bergstra & Bengio, 2012; Meaney, Wang, Guan, & Stukel, 2025).

Comparing MAEF, which is strictly aimed at minimizing the mean absolute error, with HEF, designed to balance explanatory power, cumulative accuracy, and the penalization of extreme errors, a consistent pattern was observed across all training configurations (91:9, 80:20, and 70:30) and with the three optimizers used (Grid Search, PSO, and Optuna). In each case, MAEF showed advantages in metrics focused on mean error (MAE/MASE) and runtime, while HEF concentrated on improvements in $R^2$, GRA, and RMSE/RMSSE, consolidating its resilience against large errors (Hernandez, Saini, & Moore, 2024; Vargas-Forero, Manotas-Duque, & Trujillo, 2024; Chang, Li, & Lin, 2024).

Statistical contrasts confirmed the validity of these differences: across all partitions and with the three optimizers, the difference of proportions tests yielded highly significant Z values, ruling out the influence of chance or the search method. This finding reinforces that the improvements stem from the evaluation function used, rather than from the optimizer. Furthermore, the visual analysis of comparative Figures 6, 10, and 14 supports this observation: in $R^2$ and GRA, the models optimized with HEF mostly rank above the reference diagonal, while in MAE and MASE, the best results are concentrated in MAEF. In the case of RMSE and RMSSE, the pattern was mixed, although HEF predominated in reducing extreme errors. In terms of models, HEF showed more pronounced advantages in complex architectures, such as CatBoost, XGBoost, MLP, and DNN-LSTM, whereas MAEF performed more favorably in simpler models, including SVR, KNN, and linear regression. This behavior suggests that the hierarchical integration of metrics proposed by HEF provides greater value in scenarios where model complexity amplifies the risk of overfitting or severe errors.

The analysis of different train–test partition schemes reveals that the relative advantages of HEF remain stable as the proportion of training data decreases. While the absolute performance of models naturally deteriorates when shifting from a 91:9 to a 70:30 split, HEF consistently outperforms MAEF in global metrics such as $R^2$, GRA, RMSE, and RMSSE. This behavior suggests that the hierarchical and penalty-based structure of HEF enhances generalization capacity and robustness in the optimization process, particularly in data-scarce scenarios where single-metric approaches tend to be more unstable (Petropoulos, Fildes, & Goodwin, 2016; Salinas, Flunkert, Gasthaus, & Januschowski, 2020). The metrics used in this comparison have been widely validated in the literature for their interpretability and operational relevance (Chicco, Warrens, & Jurman, 2021).

To further enhance interpretability, illustrative case studies were conducted to compare scenarios in which HEF and MAEF selected different models. For instance, in one example from the Walmart dataset, MAEF favored a linear regression due to its lower mean absolute error, whereas HEF selected a CatBoost model. Although the latter exhibited a slightly higher MAE, residual plots revealed that it produced fewer extreme errors and more stable forecasts across deciles. This practical contrast highlights the effect of the hierarchical penalization mechanism: HEF systematically avoids models that overfit global fit metrics (e.g., $R^2$) while failing to control extreme deviations, thus offering a more reliable basis for operational decision-making.

Beyond the empirical results, it is also important to position HEF within the broader landscape of model evaluation practices. In forecasting competitions such as M4 and M5, standardized error measures like sMAPE, MASE, and RMSSE have become benchmarks for comparison, but they do not incorporate adaptive thresholds or hierarchical penalization, which limits their ability to capture different dimensions of performance simultaneously. In parallel, research on hyperparameter optimization confirms the value of metaheuristic techniques, such as PSO and related approaches, in enhancing model accuracy and stability (Khan, et al., 2024). Meanwhile, studies emphasize the critical role of hyperparameter configuration in improving robustness (Arnold, Biedebach, Küpfer, & Neunhoeffer, 2024). Likewise, adaptive model selection methods, such as Lexicase, demonstrate how flexible evaluation criteria can be aligned with domain-specific objectives (Hernandez, Saini, & Moore, 2024). Recent applications of hybrid ensembles with Bayesian optimization (Iqbal, Gil, & Kim, 2025) and gradient boosting models with hyperparameter tuning (Pandove & Deepika., 2025) illustrate a growing interest in combining diverse evaluation signals, although without adopting hierarchical or penalty-based structures. Against this background, HEF makes a distinctive contribution by introducing a hierarchical and interpretable evaluation function that explicitly prioritizes explanatory power ($R^2$), mean accuracy (MAE), and resilience to extreme deviations (RMSE), complemented by adaptive penalties. This design bridges the gap between traditional single-metric benchmarks and recent optimization-driven approaches, offering a robust yet transparent alternative for demand forecasting.

Taken together, the findings reveal a methodological trade-off: MAEF guarantees computational efficiency and strict control of mean error, while HEF maximizes explanatory power, resilience, and overall forecast stability. Therefore, the choice between the two functions must be aligned with the operational objective of the application: HEF when it is necessary to prioritize the overall quality and resilience of predictions in dynamic and volatile environments, and MAEF when the priority is to reduce mean error and optimize efficiency in more stable or less complex scenarios.

A limitation of the present study is the adoption of fixed hierarchical weights ($R^2$ = 1.0, MAE = 1.0, RMSE = 0.5) within the evaluation function. These weights were selected based on evidence reported in the literature and were maintained constant across all experiments to ensure comparability and to isolate the effect of the hierarchical design. While this decision provided methodological consistency, it also constrained the exploration of alternative configurations and their potential influence on the results.

## 6. Conclusion

### *6.1. Main findings*

The study confirms that the choice of evaluation function has a decisive impact on the performance of demand prediction models. The comparison between MAEF and HEF, conducted under different training and testing partitions (91:9, 80:20, and 70:30) and with multiple optimizers, revealed a consistent and statistically significant pattern that persists across various experimental setups. In terms of performance, HEF consistently outperforms MAEF in global metrics, including $R^2$, GRA, RMSE, and RMSSE, thereby strengthening the models' explanatory power and increasing their robustness to large errors. These results are corroborated by both the case-by-case analyses (see Appendices D–F) and the visual comparisons (Figures 3, 4, 5, 7, 8, 9, 11, 12, and 13), and validated by statistical tests of difference in proportions, which yielded Z values between −33.18 and −66.21 ($p \approx 0.0$). The consistency of this evidence demonstrates that the advantages of HEF derive directly from the evaluation function, rather than from the optimizer used.

### *6.2. Methodological trade-off*

In contrast, MAEF retains clear advantages in reducing absolute errors (MAE and MASE) and in runtimes, positioning it as an efficient alternative in scenarios where computational simplicity and mean error control are priorities. This behavior was particularly evident in simpler models, such as linear regressors, SVRs, and KNNs, which benefit from an evaluation approach focused on localized errors. The cross-sectional analysis of the results thus reveals a robust methodological trade-off: HEF emerges as the most appropriate option when the objective is to maximize explanatory power, overall robustness, and cumulative accuracy over extended prediction horizons, while MAEF proves more advantageous in operational contexts where time efficiency and strict reduction of mean error are priorities.

### *6.3. Practical implications*

In summary, the choice between HEF and MAEF should not be viewed as mutually exclusive, but rather as a strategic decision that depends on the application's objectives. For business planning scenarios and long-term horizons, where stability and generalization capacity are critical, HEF constitutes the most robust alternative; while in short-term applications or in environments with time and resource constraints, MAEF represents the most efficient option.

### *6.4. Limitations and Future Work*

Despite the consistent advantages demonstrated by HEF, this study presents certain limitations that open avenues for future research. First, the hierarchical weights ($\omega R^2$, $\omega MAE$, $\omega RMSE$) and the coefficient of variation thresholds used to define tolerances were adopted from existing literature and maintained fixed throughout all experiments. While this approach ensured comparability and methodological focus on validating HEF as a functional evaluation framework, it did not explore the sensitivity of results to alternative parameterizations. Future research should therefore address this limitation by conducting ablation and sensitivity analyses that vary hierarchical weights and tolerance thresholds, in order to assess the robustness and generalizability of HEF under different configurations. Moreover, metaheuristic approaches could be investigated to dynamically adjust these weights according to the statistical properties of each dataset, thereby enhancing the adaptability and generalization capacity of HEF. Additionally, extending validation to a broader range of datasets and forecasting horizons, as well as exploring its integration into AutoML frameworks, represent promising directions for expanding the scope and applicability of the proposed function. At the same time, it is important to emphasize that HEF does not operate as a Pareto-based multi-objective optimization. Future research may therefore compare its hierarchical integration with formal multi-objective frameworks to further assess trade-offs among accuracy, robustness, and explanatory power.

Additionally, future work should expand the empirical validation of HEF to other application domains such as energy demand forecasting, healthcare resource planning, and logistics. These sectors exhibit distinct temporal dynamics and sources of non-stationarity, such as periodic shocks, irregular demand surges, or capacity constraints, that would enable a more comprehensive assessment of HEF's generalizability and robustness across diverse forecasting environments.

In this regard, future research could further explore the sensitivity of hierarchical weights and tolerance thresholds by evaluating alternative configurations or adaptive strategies based on the statistical characteristics of each series (Kourentzes, Trapero, & Barrow, 2020; Bergmeir, Hyndman, & Koo, 2018). Recent studies suggest that dynamically optimizing evaluation criteria may enhance generalization in highly heterogeneous contexts, albeit at the cost of increased computational complexity (Makridakis, Spiliotis, & Assimakopoulos, 2022; Montero-Manso & Hyndman, 2021; Criado-Ramón, Ruiz, & Pegalajar, 2025).

Furthermore, it is worth noting that percentage-based metrics, such as sMAPE and MAPE, were deliberately excluded from the formulation of HEF due to their well-documented numerical instabilities, including scale sensitivity and undefined values near zero. Although this decision ensured robustness in the present experiments, integrating such metrics into hybrid evaluation functions remains a potential line of research, particularly in domains where relative error has practical relevance. Likewise, future studies should assess the applicability of HEF in hierarchical forecasting tasks, such as multi-level demand prediction across store, regional, and national levels, where the function's ability to balance accuracy and explanatory power could provide significant advantages. These directions would allow consolidating HEF as a generalizable and adaptive evaluation framework for real-world forecasting environments.

### CRediT authorship contribution statement

**Adolfo González**: Conceptualization, Methodology, Software, Writing – original draft, Investigation, Supervision, Writing – review & editing. **Victor Parada:** Writing – review & editing.

### Declaration of competing interest

The authors declare that they have no known competing financial interests or personal relationships that could have appeared to influence the work reported in this paper.

### Funding

This research received no external funding.

### Data availability

The datasets analyzed in this study are publicly available from the following sources: Walmart dataset (Yasser, 2021), M3 (Makridakis & Hibon, The M3-Competition: results, conclusions and implications, 2000), M4 (Makridakis, Spiliotis, & Assimakopoulos, The M4 Competition: Results, findings, conclusion and way forward, 2018), and M5 (Makridakis, Spiliotis, & Assimakopoulos, M5 accuracy competition: Results, findings, and conclusions, 2022).

## Appendix A

**Table A1.** Fixed Parameters and Hyperparameters.

| Model | Parameters |
|---|---|
| ARIMA | p=1, d=1, q=1 |
| KNN | n_neighbors = 5 |
| SVR | kernel: rbf; C: 1; degree: 3; epsilon: 0.1; gamma: scale |
| LR | Defaults |
| LSR | alpha: 0.01 |
| RR | alpha: 0.01 |
| ENR | alpha=0.01; l1_ratio=0.1 |
| DTR | max_depth: None |
| GBR | n_estimators=90; learning_rate=0.1 |
| RFR | n_estimators=100 |
| XGBoost | n_estimators=700; learning_rate=0.1; max_depth=3 |
| HR | epsilon=1 |
| RPL | degree=2 |
| BR | max_iter=50; alpha_1=1e-2; alpha_2=1e-2 |
| SES | s_l: 0.2 |
| CATBOOST | iterations: 300 depth: 6 learning_rate: 0.05 |
| MLP | c1: 32; c2: 32; c3: 16 |
| LSTM | n_epochs: 50 |
| DNN-LSTM | n_epochs: 50 |

## Appendix B

**Table B1.** Execution scheme of the experimental protocol.

| ID | 1 | 2 | 3 | 4 | 5 | | | 6 | 7 | 8 | 9 |
|---|---|---|---|---|---|---|---|---|---|---|---|
| | | | | | 5(a) | 5(b) | 5(c) | | | | |
| 1 | IS | 91:9 | X | X | - | - | - | - | - | X | 21 |
| 2 | IS | 80:20 | X | X | - | - | - | - | - | X | 21 |
| 3 | IS | 70:30 | X | X | - | - | - | - | - | X | 21 |
| 4 | IS | 91:9 | X | - | X | X | - | X | - | X | 21 |
| 5 | IS | 80:20 | X | - | X | X | - | X | - | X | 21 |
| 6 | IS | 70:30 | X | - | X | X | - | X | - | X | 21 |
| 7 | IS | 91:9 | X | - | X | - | X | X | - | X | 21 |
| 8 | IS | 80:20 | X | - | X | - | X | X | - | X | 21 |
| 9 | IS | 70:30 | X | - | X | - | X | X | - | X | 21 |
| 10 | IS | 91:9 | X | - | X | X | - | - | X | X | 21 |
| 11 | IS | 80:20 | X | - | X | X | - | - | X | X | 21 |
| 12 | IS | 70:30 | X | - | X | X | - | - | X | X | 21 |
| 13 | IS | 91:9 | X | - | X | - | X | - | X | X | 21 |
| 14 | IS | 80:20 | X | - | X | - | X | - | X | X | 21 |
| 15 | IS | 70:30 | X | - | X | - | X | - | X | X | 21 |

## Appendix C

*Experimental Environment, Optimizer Configurations, and Parameter Search Spaces*

The computational experiments were executed in Jupyter Notebooks within a Conda environment using Python 3.12.7, running on macOS (Darwin Kernel 24.6.0) on a 64-bit ARM processor (Apple M4) with 14 physical cores, 14 logical cores, and 32 GB of RAM. All runs were performed in CPU mode to ensure reproducibility. Although the processor includes a GPU, no GPU-specific acceleration frameworks (e.g., CUDA or TensorFlow-Metal) were invoked.

*Optimizer configurations*

- Optuna: Each training and optimization cycle was performed under a fixed budget of 24 trials.
- PSO: Configured with three particles in the swarm, an inertia weight of 0.1, a cognitive weight of 0.2 (personal best), a social weight of 0.3 (global best), and a maximum of seven iterations.

- Grid Search: Applied through exhaustive parameter sweeps, systematically exploring all possible combinations in the defined spaces.

*Parameter search spaces*

Table C1 reports the parameter search spaces explored for each model during training and optimization. These ranges were defined based on prior studies and the experimental design to ensure comparability across models.

**Table C1.** Parameter ranges explored for the models considered in this study.

| Model | Parameters |
|---|---|
| ARIMA | p=0 between 5, d=0 between 3, q=0 between 5 |
| KNN | n_neighbors between 1 and 35 |
| SVR | kernel = 'rbf'; C between 0.01 and 20; degree between 2 and 5; epsilon between 0.01 and 1; gamma = 'scale'. |
| LR | Defaults |
| LSR | alpha between 0.01 and 5; random_state = seed_value; max_iter = 8000. |
| RR | alpha between 0.1 and 10; max_iter = 8000 |
| ENR | alpha between 0.01 and 0.02; l1_ratio between 1e-3 and 0.1 |
| DTR | max_depth between 1 and 32; min_samples_split = 2; min_samples_leaf = 1; random_state = 21 |
| GBR | n_estimators between 10 and 2000; learning_rate between 0.1 and 0.3; random_state = seed_value |
| RFR | n_estimators between 1 and 100; random_state = 21 |
| XGBoost | n_estimators between 10 and 2000; learning_rate between 0.1 and 0.3; max_depth between 3 and 5; enable_categorical = True; random_state = seed_value |
| HR | epsilon between 1 and 3; max_iter = 1000; alpha = 0.0001; tol = 1e-4 |
| RPL | degree between 1 and 3; include_bias = False; interaction_only = True |
| BR | max_iter between 50 and 1000; alpha_1 between 1e-6 and 1e-2; alpha_2 between 1e-6 and 1e-2 |
| SES | smoothing_level between 0.01 and 0.9; optimized = False |
| CATBOOST | iterations between 100 and 1000; depth between 3 and 10; learning_rate between 0.01 and 0.3; loss_function = 'Huber:delta=1.0'; train_dir = tmp_dir; verbose = False; random_seed = 21 |
| MLP | hidden_layer_sizes = [layer 1: 16–32; layer 2: 0–32; layer 3: 0–32]; activation = 'relu'; solver = 'adam'; max_iter = 1000; early_stopping = True; n_iter_no_change = 10; random_state = 21 |
| LSTM | epochs between 1 and 80; batch_size = 32 |
| DNN-LSTM | epochs between 1 and 70; batch_size = 32 |

# Appendix D

*Detailed Results per Dataset - Training 91% and Testing 9%*

**Table D1.** Comparison in number of cases with Grid Search.

| Metrics | MAEF versus the non-optimized baseline | | | HEF versus the baseline | | | HEF versus MAEF | | |
|---|---|---|---|---|---|---|---|---|---|
| | Improves MAEF | Improves Base | No Change | Improves HEF | Improves Base | No Change | Improves HEF | Improves MAEF | No Change |
| $R^2$ | **4099** | 0 | 5379 | **4530** | 0 | 4948 | **1673** | 0 | 7805 |
| MAE | 1 | **2088** | 7389 | 1 | **2294** | 7183 | 0 | **1386** | 8092 |
| RMSE | 0 | **2804** | 6674 | 0 | **2461** | 7017 | 0 | **514** | 8964 |
| RMSSE | 1 | **2494** | 6983 | 1 | **2089** | 7388 | 0 | **424** | 9054 |
| MASE | 0 | **1804** | 7674 | 0 | **2051** | 7427 | 0 | **1458** | 8020 |
| GRA | **3887** | 0 | 5591 | **4145** | 0 | 5333 | **1281** | 0 | 8197 |
| Execution Time | 74 | **8864** | 540 | 90 | **8893** | 495 | 386 | **4574** | 4518 |

**Table D2.** Comparison in number of cases optimized with PSO.

| Metrics | MAEF versus the non-optimized baseline | | | HEF versus the baseline | | | HEF versus MAEF | | |
|---|---|---|---|---|---|---|---|---|---|
| | Improves MAEF | Improves Base | No Change | Improves HEF | Improves Base | No Change | Improves HEF | Improves MAEF | No Change |
| $R^2$ | **5274** | 3893 | 5819 | **5608** | 3620 | 5758 | **1639** | 288 | 13059 |
| MAE | **5760** | 3923 | 5303 | **5537** | 4074 | 5375 | 270 | **1017** | 13699 |
| RMSE | **5235** | 4348 | 5403 | **5521** | 4199 | 5266 | **969** | 235 | 13782 |
| RMSSE | **5492** | 3912 | 5582 | **5831** | 3779 | 5376 | **1230** | 272 | 13484 |
| MASE | **5993** | 3803 | 5190 | **5793** | 3943 | 5250 | 238 | **1073** | 13675 |
| GRA | **4962** | 3851 | 6173 | **5235** | 3621 | 6130 | **1207** | 429 | 13350 |
| Execution Time | 72 | **14422** | 492 | 90 | **14402** | 494 | 480 | **2723** | 11783 |

**Table D3.** Comparison in number of cases optimized with Optuna.

| Metrics | MAEF versus the non-optimized baseline | | | HEF versus the baseline | | | HEF versus MAEF | | |
|---|---|---|---|---|---|---|---|---|---|
| | Improves MAEF | Improves Base | No Change | Improves HEF | Improves Base | No Change | Improves HEF | Improves MAEF | No Change |
| $R^2$ | **6153** | 3297 | 5536 | **7232** | 2811 | 4943 | **2518** | 274 | 12194 |
| MAE | **6571** | 3105 | 5310 | **6067** | 3578 | 5341 | 248 | **1685** | 13053 |
| RMSE | **5387** | 3862 | 5737 | **6002** | 3660 | 5324 | **1398** | 360 | 13228 |
| RMSSE | **5941** | 3551 | 5494 | **6672** | 3206 | 5108 | **1766** | 324 | 12896 |
| MASE | **7065** | 3073 | 4848 | **6551** | 3535 | 4900 | 279 | **1799** | 12908 |
| GRA | **5356** | 3219 | 6411 | **5842** | 2776 | 6368 | **1869** | 538 | 12579 |
| Execution Time | 6 | **14673** | 307 | 0 | **14707** | 279 | 1502 | **2464** | 11020 |

**Table D4.** Dataset Walmart: Comparison in number of cases.

| Metrics | Models optimized with Grid Search and HEF versus those optimized with Grid Search and MAEF | | | Models optimized with PSO and HEF versus those optimized with PSO and MAEF | | | Models optimized with Optuna and HEF versus those optimized with Optuna and MAEF | | |
|---|---|---|---|---|---|---|---|---|---|
| | Improves HEF | Improves MAEF | No Change | Improves HEF | Improves MAEF | No Change | Improves HEF | Improves MAEF | No Change |
| $R^2$ | **74** | 0 | 248 | **76** | 30 | 492 | **111** | 15 | 472 |
| MAE | 0 | **61** | 261 | 17 | **29** | 552 | 15 | **55** | 528 |
| RMSE | 0 | **24** | 298 | **46** | 15 | 537 | **63** | 18 | 517 |
| RMSSE | 0 | **21** | 301 | **60** | 41 | 497 | **107** | 25 | 466 |
| MASE | 0 | **64** | 258 | 38 | **65** | 495 | 19 | **94** | 485 |
| GRA | **47** | 0 | 275 | **56** | 23 | 519 | **65** | 24 | 509 |
| Execution Time | 23 | **117** | 182 | 32 | **104** | 462 | 23 | **137** | 438 |

**Table D5.** Dataset M3: Comparison in number of cases.

| Metrics | Models optimized with Grid Search and HEF versus those optimized with Grid Search and MAEF | | | Models optimized with PSO and HEF versus those optimized with PSO and MAEF | | | Models optimized with Optuna and HEF versus those optimized with Optuna and MAEF | | |
|---|---|---|---|---|---|---|---|---|---|
| | Improves HEF | Improves MAEF | No Change | Improves HEF | Improves MAEF | No Change | Improves HEF | Improves MAEF | No Change |
| $R^2$ | **412** | 0 | 2766 | **348** | 84 | 4562 | **526** | 83 | 4385 |
| MAE | 0 | **300** | 2878 | 100 | **194** | 4700 | 100 | **259** | 4635 |
| RMSE | 0 | **151** | 3027 | **194** | 85 | 4715 | **297** | 100 | 4597 |
| RMSSE | 0 | **115** | 3063 | **280** | 90 | 4624 | **401** | 89 | 4504 |
| MASE | 0 | **335** | 2843 | 83 | **237** | 4674 | 105 | **397** | 4492 |
| GRA | **237** | 0 | 2941 | **198** | 126 | 4670 | **298** | 153 | 4543 |
| Execution Time | 111 | **1418** | 1649 | 164 | **867** | 3963 | 452 | **632** | 3910 |

**Table D6.** Dataset M4: Comparison in number of cases.

| Metrics | Models optimized with Grid Search and HEF versus those optimized with Grid Search and MAEF | | | Models optimized with PSO and HEF versus those optimized with PSO and MAEF | | | Models optimized with Optuna and HEF versus those optimized with Optuna and MAEF | | |
|---|---|---|---|---|---|---|---|---|---|
| | Improves HEF | Improves MAEF | No Change | Improves HEF | Improves MAEF | No Change | Improves HEF | Improves MAEF | No Change |
| $R^2$ | **189** | 0 | 1239 | **129** | 52 | 2063 | **207** | 45 | 1992 |
| MAE | 0 | **157** | 1271 | 47 | **85** | 2112 | 34 | **126** | 2084 |
| RMSE | 0 | **65** | 1363 | **88** | 30 | 2126 | **126** | 42 | 2076 |
| RMSSE | 0 | **60** | 1368 | **92** | 26 | 2126 | **175** | 39 | 2030 |
| MASE | 0 | **170** | 1258 | 44 | **94** | 2106 | 44 | **179** | 2021 |
| GRA | **126** | 0 | 1302 | **68** | 64 | 2112 | **122** | 82 | 2040 |
| Execution Time | 141 | **534** | 753 | 55 | **492** | 1697 | **302** | 90 | 1852 |

**Table D7.** Dataset M5: Comparison in number of cases.

| Metrics | Models optimized with Grid Search and HEF versus those optimized with Grid Search and MAEF | | | Models optimized with PSO and HEF versus those optimized with PSO and MAEF | | | Models optimized with Optuna and HEF versus those optimized with Optuna and MAEF | | |
|---|---|---|---|---|---|---|---|---|---|
| | Improves HEF | Improves MAEF | No Change | Improves HEF | Improves MAEF | No Change | Improves HEF | Improves MAEF | No Change |
| $R^2$ | **998** | 0 | 3552 | **1086** | 122 | 5942 | **1674** | 131 | 5345 |
| MAE | 0 | **868** | 3682 | 106 | **709** | 6335 | 99 | **1245** | 5806 |
| RMSE | 0 | **274** | 4276 | **641** | 105 | 6404 | **912** | 200 | 6038 |
| RMSSE | 0 | **228** | 4322 | **798** | 115 | 6237 | **1083** | 171 | 5896 |
| MASE | 0 | **889** | 3661 | 73 | **677** | 6400 | 111 | **1129** | 5910 |
| GRA | **871** | 0 | 3679 | **885** | 216 | 6049 | **1384** | 279 | 5487 |
| Execution Time | 111 | **2505** | 1934 | 229 | **1260** | 5661 | 725 | **1605** | 4820 |

**Table D8.** Statistical comparison using the difference of proportions test (Z, p) under the 91:9 training/testing split for HEF vs. MAEF across datasets.

| Dataset | Grid Search (Z, p) | PSO (Z, p) | Optuna (Z, p) |
|---|---|---|---|
| Walmart | −6.10 ; 1.07e−09 | −7.19 ; 6.44e−13 | −13.37 ; 8.68e−41 |
| M3 | −12.90 ; 4.75e−38 | −17.25 ; 1.18e−66 | −25.49 ; 2.53e−143 |
| M4 | −9.24 ; 2.52e−20 | −8.89 ; 6.33e−19 | −14.93 ; 2.09e−50 |
| M5 | −29.04 ; 2.35e−185 | −46.93 ; 0.0 | −59.02 ; 0.0 |

# Appendix E

*Detailed Results per Dataset - Training 80% and Testing 20%*

**Table E1.** Comparison in number of cases with Grid Search.

| Metrics | MAEF versus the non-optimized baseline | | | HEF versus the baseline | | | HEF versus MAEF | | |
|---|---|---|---|---|---|---|---|---|---|
| | Improves MAEF | Improves Base | No Change | Improves HEF | Improves Base | No Change | Improves HEF | Improves MAEF | No Change |
| $R^2$ | **4573** | 0 | 5577 | **4790** | 0 | 4688 | **1898** | 0 | 7580 |
| MAE | 0 | **2243** | 7907 | 0 | **2452** | 7026 | 0 | **1670** | 7808 |
| RMSE | 0 | **3026** | 7124 | 0 | **2469** | 7009 | 0 | **624** | 8854 |
| RMSSE | 0 | **2840** | 7310 | 0 | **2247** | 7231 | 0 | **537** | 8941 |
| MASE | 0 | **2024** | 8126 | 0 | **2217** | 7261 | 0 | **1733** | 7745 |
| GRA | **4074** | 0 | 6076 | **4150** | 0 | 5328 | **1448** | 0 | 8030 |
| Execution Time | 68 | **9452** | 630 | 24 | **8891** | 563 | 523 | **4040** | 4915 |

**Table E2.** Comparison in number of cases optimized with PSO.

| Metrics | MAEF versus the non-optimized baseline | | | HEF versus the baseline | | | HEF versus MAEF | | |
|---|---|---|---|---|---|---|---|---|---|
| | Improves MAEF | Improves Base | No Change | Improves HEF | Improves Base | No Change | Improves HEF | Improves MAEF | No Change |
| $R^2$ | **5024** | 4085 | 5877 | **5436** | 3755 | 5795 | **1794** | 221 | 12971 |
| MAE | **5538** | 4080 | 5368 | **5311** | 4246 | 5429 | 247 | **1163** | 13576 |
| RMSE | **5159** | 4398 | 5429 | **5497** | 4157 | 5332 | **1205** | 230 | 13551 |
| RMSSE | **5248** | 4088 | 5650 | **5636** | 3862 | 5488 | **1394** | 237 | 13355 |
| MASE | **5761** | 3995 | 5230 | **5474** | 4062 | 5450 | 248 | **1194** | 13544 |
| GRA | **4817** | 4037 | 6132 | **4987** | 3804 | 6195 | **1148** | 634 | 13204 |
| Execution Time | 51 | **14389** | 546 | 23 | **14429** | 534 | 549 | **2364** | 12073 |

**Table E3.** Comparison in number of cases optimized with Optuna.

| Metrics | MAEF versus the non-optimized baseline | | | HEF versus the baseline | | | HEF versus MAEF | | |
|---|---|---|---|---|---|---|---|---|---|
| | Improves MAEF | Improves Base | No Change | Improves HEF | Improves Base | No Change | Improves HEF | Improves MAEF | No Change |
| $R^2$ | **6369** | 3299 | 5318 | **7463** | 2796 | 4727 | **2718** | 289 | 11979 |
| MAE | **6430** | 3158 | 5398 | **5903** | 3713 | 5370 | 254 | **1862** | 12870 |
| RMSE | **5629** | 3787 | 5570 | **6242** | 3442 | 5302 | **1574** | 361 | 13051 |
| RMSSE | **6097** | 3598 | 5291 | **6803** | 3237 | 4946 | **1894** | 370 | 12722 |
| MASE | **6883** | 3289 | 4814 | **6247** | 3746 | 4993 | 275 | **1931** | 12780 |
| GRA | **5198** | 3594 | 6194 | **5587** | 3318 | 6081 | **1692** | 914 | 12380 |
| Execution Time | 16 | **14511** | 459 | 23 | **14481** | 482 | 1264 | **2599** | 11123 |

**Table E4.** Dataset Walmart: Comparison in number of cases.

| Metrics | Models optimized with Grid Search and HEF versus those optimized with Grid Search and MAEF | | | Models optimized with PSO and HEF versus those optimized with PSO and MAEF | | | Models optimized with Optuna and HEF versus those optimized with Optuna and MAEF | | |
|---|---|---|---|---|---|---|---|---|---|
| | Improves HEF | Improves MAEF | No Change | Improves HEF | Improves MAEF | No Change | Improves HEF | Improves MAEF | No Change |
| $R^2$ | **66** | 0 | 256 | **51** | 16 | 531 | **79** | 29 | 490 |
| MAE | 0 | **51** | 271 | 20 | **21** | 557 | 20 | **57** | 521 |
| RMSE | 0 | **16** | 306 | **27** | 16 | 555 | **51** | 23 | 524 |
| RMSSE | 0 | **12** | 310 | **54** | 19 | 525 | **66** | 41 | 491 |
| MASE | 0 | **64** | 258 | **39** | 33 | 526 | 20 | **79** | 499 |
| GRA | **36** | 0 | 286 | **30** | 22 | 546 | **56** | 31 | 511 |
| Execution Time | 44 | **57** | 221 | **68** | 35 | 495 | 17 | **111** | 470 |

**Table E5.** Dataset M3: Comparison in number of cases.

| Metrics | Models optimized with Grid Search and HEF versus those optimized with Grid Search and MAEF | | | Models optimized with PSO and HEF versus those optimized with PSO and MAEF | | | Models optimized with Optuna and HEF versus those optimized with Optuna and MAEF | | |
|---|---|---|---|---|---|---|---|---|---|
| | Improves HEF | Improves MAEF | No Change | Improves HEF | Improves MAEF | No Change | Improves HEF | Improves MAEF | No Change |
| $R^2$ | **480** | 0 | 2698 | **372** | 89 | 4533 | **596** | 88 | 4310 |
| MAE | 0 | **435** | 2743 | 105 | **263** | 4626 | 91 | **391** | 4512 |
| RMSE | 0 | **191** | 2987 | **319** | 87 | 4588 | **380** | 83 | 4531 |
| RMSSE | 0 | **149** | 3029 | **324** | 98 | 4572 | **461** | 96 | 4437 |
| MASE | 0 | **459** | 2719 | 75 | **354** | 4565 | 82 | **501** | 4411 |
| GRA | **309** | 0 | 2869 | **231** | 204 | 4559 | 295 | **302** | 4397 |
| Execution Time | 74 | **913** | 2191 | 149 | **842** | 4003 | 251 | **740** | 4003 |

**Table E6.** Dataset M4: Comparison in number of cases.

| Metrics | Models optimized with Grid Search and HEF versus those optimized with Grid Search and MAEF | | | Models optimized with PSO and HEF versus those optimized with PSO and MAEF | | | Models optimized with Optuna and HEF versus those optimized with Optuna and MAEF | | |
|---|---|---|---|---|---|---|---|---|---|
| | Improves HEF | Improves MAEF | No Change | Improves HEF | Improves MAEF | No Change | Improves HEF | Improves MAEF | No Change |
| $R^2$ | **218** | 0 | 1210 | **135** | 40 | 2069 | **242** | 37 | 1965 |
| MAE | 0 | **202** | 1226 | 44 | **138** | 2062 | 33 | **167** | 2044 |
| RMSE | 0 | **89** | 1339 | **153** | 30 | 2061 | **161** | 40 | 2043 |
| RMSSE | 0 | **89** | 1339 | **113** | 41 | 2090 | **180** | 41 | 2023 |
| MASE | 0 | **209** | 1219 | 41 | **112** | 2091 | 46 | **160** | 2038 |
| GRA | **142** | 0 | 1286 | 77 | **137** | 2030 | 107 | **137** | 2000 |
| Execution Time | 288 | **610** | 530 | 52 | **451** | 1741 | 144 | **271** | 1829 |

**Table E7.** Dataset M5: Comparison in number of cases.

| Metrics | Models optimized with Grid Search and HEF versus those optimized with Grid Search and MAEF | | | Models optimized with PSO and HEF versus those optimized with PSO and MAEF | | | Models optimized with Optuna and HEF versus those optimized with Optuna and MAEF | | |
|---|---|---|---|---|---|---|---|---|---|
| | Improves HEF | Improves MAEF | No Change | Improves HEF | Improves MAEF | No Change | Improves HEF | Improves MAEF | No Change |
| $R^2$ | **1134** | 0 | 3416 | **1236** | 76 | 5838 | **1801** | 135 | 5214 |
| MAE | 0 | **982** | 3568 | 78 | **741** | 6331 | 110 | **1247** | 5793 |
| RMSE | 0 | **328** | 4222 | **706** | 97 | 6347 | **982** | 215 | 5953 |
| RMSSE | 0 | **287** | 4263 | **903** | 79 | 6168 | **1187** | 192 | 5771 |
| MASE | 0 | **1001** | 3549 | 93 | **695** | 6362 | 127 | **1191** | 5832 |
| GRA | **961** | 0 | 3589 | **810** | 271 | 6069 | **1234** | 444 | 5472 |
| Execution Time | 117 | **2460** | 1973 | 280 | **1036** | 5834 | 852 | **1477** | 4821 |

**Table E8.** Statistical comparison using the difference of proportions test (Z, p) under the 80:20 training/testing split for HEF vs. MAEF across datasets.

| Dataset | Grid Search (Z, p) | PSO (Z, p) | Optuna (Z, p) |
|---|---|---|---|
| Walmart | −6.66 ; 2.73e−11 | −5.95 ; 2.62e−09 | −6.88 ; 6.12e−12 |
| M3 | −13.67 ; 1.54e−42 | −18.91 ; 9.60e−80 | −24.97 ; 1.15e−137 |
| M4 | −8.04 ; 9.12e−16 | −8.71 ; 2.93e−18 | −14.54 ; 6.92e−48 |
| M5 | −29.55 ; 6.31e−192 | −50.33 ; 0.0 | −56.77 ; 0.0 |

# Appendix F

*Detailed Results per Dataset - Training 70% and Testing 30%*

**Table F1.** Comparison in number of cases with Grid Search.

| Metrics | MAEF versus the non-optimized baseline | | | HEF versus the baseline | | | HEF versus MAEF | | |
|---|---|---|---|---|---|---|---|---|---|
| | Improves MAEF | Improves Base | No Change | Improves HEF | Improves Base | No Change | Improves HEF | Improves MAEF | No Change |
| $R^2$ | **4371** | 0 | 5107 | **4837** | 0 | 4641 | **2035** | 0 | 7443 |
| MAE | 0 | **2346** | 7132 | 0 | **2653** | 6825 | 0 | **1744** | 7734 |
| RMSE | 0 | **2763** | 6715 | 0 | **2363** | 7115 | 0 | **651** | 8827 |
| RMSSE | 0 | **2574** | 6904 | 0 | **2155** | 7323 | 0 | **574** | 8904 |
| MASE | 0 | **1942** | 7536 | 0 | **2267** | 7211 | 0 | **1838** | 7640 |
| GRA | **3956** | 0 | 5522 | **4245** | 0 | 5233 | **1512** | 0 | 7966 |
| Execution Time | 124 | **8618** | 736 | 113 | **8660** | 705 | 1314 | **3384** | 4780 |

**Table F2.** Comparison in number of cases optimized with PSO.

| Metrics | MAEF versus the non-optimized baseline | | | HEF versus the baseline | | | HEF versus MAEF | | |
|---|---|---|---|---|---|---|---|---|---|
| | Improves MAEF | Improves Base | No Change | Improves HEF | Improves Base | No Change | Improves HEF | Improves MAEF | No Change |
| R² | **5264** | 3873 | 5849 | **5618** | 3510 | 5858 | **1840** | 248 | 12898 |
| MAE | **5702** | 3930 | 5354 | **5423** | 4150 | 5413 | 235 | **1152** | 13599 |
| RMSE | **5129** | 4370 | 5487 | **5454** | 4142 | 5390 | **1145** | 240 | 13601 |
| RMSSE | **5369** | 4014 | 5603 | **5746** | 3707 | 5533 | **1355** | 249 | 13382 |
| MASE | **5791** | 3883 | 5312 | **5541** | 4057 | 5388 | 242 | **1151** | 13593 |
| GRA | **4923** | 3993 | 6070 | **4942** | 3690 | 6354 | **1140** | 621 | 13225 |
| Execution Time | 116 | **14172** | 698 | 113 | **14170** | 703 | 599 | **2074** | 12313 |

**Table F3.** Comparison in number of cases optimized with Optuna.

| Metrics | MAEF versus the non-optimized baseline | | | HEF versus the baseline | | | HEF versus MAEF | | |
|---|---|---|---|---|---|---|---|---|---|
| | Improves MAEF | Improves Base | No Change | Improves HEF | Improves Base | No Change | Improves HEF | Improves MAEF | No Change |
| R² | **6616** | 3109 | 5261 | **7833** | 2589 | 4564 | **2709** | 254 | 12023 |
| MAE | **6565** | 3041 | 5380 | **5973** | 3585 | 5428 | 249 | **1871** | 12866 |
| RMSE | **5595** | 3803 | 5588 | **6251** | 3529 | 5206 | **1553** | 345 | 13088 |
| RMSSE | **6100** | 3497 | 5389 | **6916** | 3121 | 4949 | **1923** | 309 | 12754 |
| MASE | **6861** | 3239 | 4886 | **6289** | 3741 | 4956 | 253 | **1827** | 12906 |
| GRA | **5173** | 3653 | 6160 | **5576** | 3401 | 6009 | **1725** | 878 | 12383 |
| Execution Time | 28 | **14361** | 597 | 23 | **14385** | 578 | 1175 | **2945** | 10866 |

**Table F4.** Dataset Walmart: Comparison in number of cases.

| Metrics | Models optimized with Grid Search and HEF versus those optimized with Grid Search and MAEF | | | Models optimized with PSO and HEF versus those optimized with PSO and MAEF | | | Models optimized with Optuna and HEF versus those optimized with Optuna and MAEF | | |
|---|---|---|---|---|---|---|---|---|---|
| | Improves HEF | Improves MAEF | No Change | Improves HEF | Improves MAEF | No Change | Improves HEF | Improves MAEF | No Change |
| R² | **64** | 0 | 258 | **66** | 29 | 503 | **81** | 22 | 495 |
| MAE | 0 | **57** | 265 | 23 | **36** | 539 | 13 | **55** | 530 |
| RMSE | 0 | **25** | 297 | **33** | 16 | 549 | **45** | 18 | 535 |
| RMSSE | 0 | **17** | 305 | **54** | 36 | 508 | **76** | 30 | 492 |
| MASE | 0 | **62** | 260 | 26 | **63** | 509 | 28 | **70** | 500 |
| GRA | **36** | 0 | 286 | 28 | **38** | 532 | **48** | 29 | 521 |
| Execution Time | 60 | **92** | 170 | 33 | **78** | 487 | 17 | **108** | 473 |

**Table F5.** Dataset M3: Comparison in number of cases.

| Metrics | Models optimized with Grid Search and HEF versus those optimized with Grid Search and MAEF | | | Models optimized with PSO and HEF versus those optimized with PSO and MAEF | | | Models optimized with Optuna and HEF versus those optimized with Optuna and MAEF | | |
|---|---|---|---|---|---|---|---|---|---|
| | Improves HEF | Improves MAEF | No Change | Improves HEF | Improves MAEF | No Change | Improves HEF | Improves MAEF | No Change |
| R² | **550** | 0 | 2628 | **440** | 77 | 4477 | **644** | 70 | 4280 |
| MAE | 0 | **426** | 2752 | 85 | **292** | 4617 | 97 | **435** | 4462 |
| RMSE | 0 | **200** | 2978 | **304** | 81 | 4609 | **407** | 85 | 4502 |
| RMSSE | 0 | **174** | 3004 | **347** | 77 | 4570 | **508** | 79 | 4407 |
| MASE | 0 | **510** | 2668 | 83 | **331** | 4580 | 78 | **464** | 4452 |
| GRA | **345** | 0 | 2833 | 202 | **244** | 4548 | **336** | 318 | 4340 |
| Execution Time | 295 | **1087** | 1796 | 218 | **624** | 4152 | 202 | **620** | 4172 |

**Table F6.** Dataset M4: Comparison in number of cases.

| Metrics | Models optimized with Grid Search and HEF versus those optimized with Grid Search and MAEF | | | Models optimized with PSO and HEF versus those optimized with PSO and MAEF | | | Models optimized with Optuna and HEF versus those optimized with Optuna and MAEF | | |
|---|---|---|---|---|---|---|---|---|---|
| | Improves HEF | Improves MAEF | No Change | Improves HEF | Improves MAEF | No Change | Improves HEF | Improves MAEF | No Change |
| R² | **241** | 0 | 1187 | **142** | 46 | 2056 | **206** | 30 | 2008 |
| MAE | 0 | **221** | 1207 | 42 | **102** | 2100 | 42 | **138** | 2064 |
| RMSE | 0 | **89** | 1339 | **128** | 41 | 2075 | **139** | 35 | 2070 |
| RMSSE | 0 | **78** | 1350 | **87** | 44 | 2113 | **150** | 34 | 2060 |
| MASE | 0 | **224** | 1204 | 47 | **92** | 2105 | 28 | **146** | 2070 |
| GRA | **152** | 0 | 1276 | **94** | 74 | 2076 | **112** | 111 | 2021 |
| Execution Time | 245 | **437** | 746 | 124 | **174** | 1946 | 44 | **642** | 1558 |

**Table F7.** Dataset M5: Comparison in number of cases.

| Metrics | Models optimized with Grid Search and HEF versus those optimized with Grid Search and MAEF | Models optimized with PSO and HEF versus those optimized with PSO and MAEF | Models optimized with Optuna and HEF versus those optimized with Optuna and MAEF |
|---|---|---|---|

| | Improves HEF | Improves MAEF | No Change | Improves HEF | Improves MAEF | No Change | Improves HEF | Improves MAEF | No Change |
|---|---|---|---|---|---|---|---|---|---|
| $R^2$ | **1180** | 0 | 3370 | **1192** | 96 | 5862 | **1778** | 132 | 5240 |
| MAE | 0 | **1040** | 3510 | 85 | **722** | 6343 | 97 | **1243** | 5810 |
| RMSE | 0 | **337** | 4213 | **680** | 102 | 6368 | **962** | 207 | 5981 |
| RMSSE | 0 | **305** | 4245 | **867** | 92 | 6191 | **1189** | 166 | 5795 |
| MASE | 0 | **1042** | 3508 | 86 | **665** | 6399 | 119 | **1147** | 5884 |
| GRA | **979** | 0 | 3571 | **816** | 265 | 6069 | **1229** | 420 | 5501 |
| Execution Time | 714 | **1768** | 2068 | 224 | **1198** | 5728 | 912 | **1575** | 4663 |

**Table F8.** Statistical comparison using the difference of proportions test (Z, p) under the 70:30 training/testing split for HEF vs. MAEF across datasets.

| Dataset | Grid Search (Z, p) | PSO (Z, p) | Optuna (Z, p) |
|---|---|---|---|
| Walmart | −5.01 ; 5.52e−07 | −3.70 ; 2.18e−04 | −8.39 ; 4.66e−17 |
| M3 | −15.00 ; 6.85e−51 | −23.44 ; 1.08e−121 | −24.97 ; 1.15e−137 |
| M4 | −9.79 ; 1.20e−22 | −9.79 ; 1.30e−22 | −14.22 ; 7.23e−46 |
| M5 | −29.55 ; 6.31e−192 | −50.33 ; 0.0 | −56.77 ; 0.0 |